\documentclass{article}

\usepackage[margin=1in]{geometry}

\usepackage{mathpazo}
\usepackage[numbers]{natbib}

\usepackage[utf8]{inputenc} 
\usepackage[T1]{fontenc}    
\usepackage[colorlinks]{hyperref}       
\usepackage{url}            
\usepackage{booktabs}       
\usepackage{amsfonts}       
\usepackage{nicefrac}       
\usepackage{microtype}      
\usepackage[table]{xcolor}         

\usepackage{color}
\usepackage{caption}
\usepackage{makecell}
\usepackage{wrapfig}
\usepackage{textcomp}
\usepackage{gensymb}
\usepackage{marginnote}
\usepackage{algorithm}
\usepackage{algorithmic}
\usepackage{enumitem}

\usepackage{tcolorbox}

\definecolor{Gray}{gray}{0.9}
\definecolor{BrickRed}{rgb}{0.6,0,0}
\definecolor{RoyalBlue}{rgb}{0,0,0.8}
\definecolor{Tdgreen}{rgb}{0,0.4,0.7}
\hypersetup{citecolor=BrickRed, linkcolor=RoyalBlue}

\usepackage{bm}
\usepackage{dsfont}
\usepackage{multirow}
\usepackage{array}
\usepackage{bbding}
\usepackage{lipsum}
\usepackage{verbatim}

\usepackage{pifont}

\usepackage{microtype}
\usepackage{graphicx}
\usepackage{subfigure}
\usepackage{booktabs} 

\usepackage{amsmath}
\usepackage{amssymb}
\usepackage{mathtools}
\usepackage{amsthm}

\usepackage[capitalize,noabbrev]{cleveref}

\theoremstyle{plain}
\newtheorem{theorem}{Theorem}[section]

\theoremstyle{definition}
\newtheorem{definition}[theorem]{Definition}

\theoremstyle{remark}

\newcommand{\setParDis}{\setlength {\parskip} {0.1cm} }

\title{Average of Pruning: Improving Performance and Stability of Out-of-Distribution Detection}

\author{
Zhen Cheng\textsuperscript{\rm 1,2}, Fei Zhu\textsuperscript{\rm 1,2}, Xu-Yao Zhang\textsuperscript{\rm 1,2}, Cheng-Lin Liu\textsuperscript{\rm 1,2}\\
\\
\normalsize \textsuperscript{\rm 1} MAIS, Institute of Automation, Chinese Academy of Sciences, Beijing 100190, China\\
\normalsize \textsuperscript{\rm 2} School of Artificial Intelligence, University of Chinese Academy of Sciences, Beijing, 100049, China\\
}
\date{}

\begin{document}

\maketitle

\begin{abstract}
Detecting Out-of-distribution (OOD) inputs have been a critical issue for neural networks in the open world. However, the unstable behavior of OOD detection along the optimization trajectory during training has not been explored clearly. In this paper, we first find the performance of OOD detection suffers from overfitting and instability during training: 1) the performance could decrease when the training error is near zero, and 2) the performance would vary sharply in the final stage of training. Based on our findings, we propose Average of Pruning (AoP), consisting of model averaging and pruning, to mitigate the unstable behaviors. 
Specifically, model averaging can help achieve a stable performance by smoothing the landscape, and pruning is certified to eliminate the overfitting by eliminating redundant features. Comprehensive experiments on various datasets and architectures are conducted to verify the effectiveness of our method.
\end{abstract}

\section{Introduction}

\label{sec:intro}

It has been found that deep neural networks would produce overconfident predictions for Out-Of-Distribution (OOD) inputs which not belong to known categories~\cite{Nguyen_2015_DeepNN}. Such a serious vulnerability to OOD inputs is likely to bring potential risks in safety-critical scenarios like autonomous driving~\cite{filos_2020_autonomous}. For example, the object detection model could possibly classify an unseen animal in training data as a traffic sign. Detecting OOD inputs has become an essential issue in practical deployment.

\begin{figure}[t]
	\begin{center}
		\includegraphics[width=0.70\linewidth]{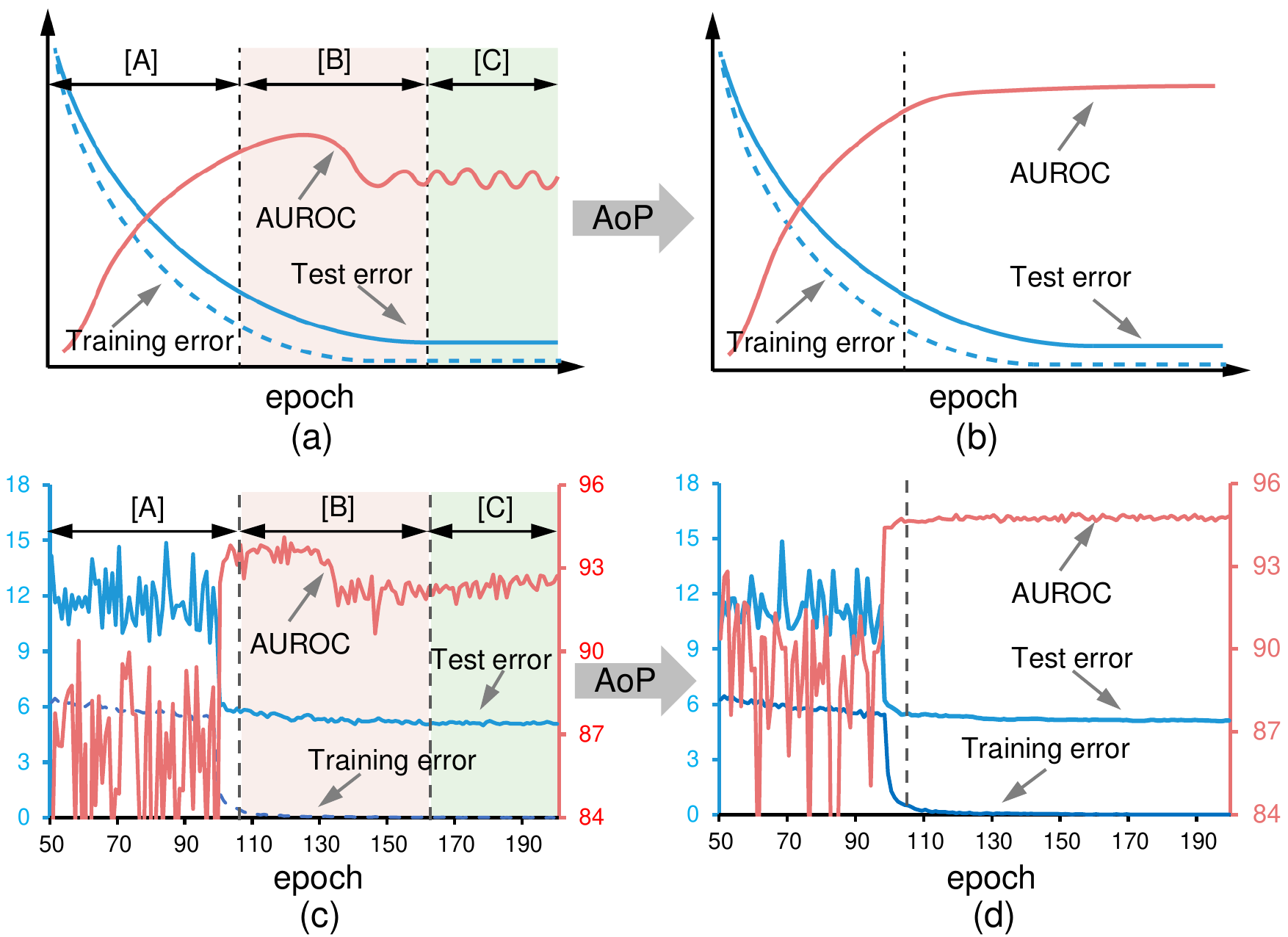}
	\end{center}
	\vskip -0.15in
	\caption{(a) shows the concepts of overfitting and instability, which harms OOD detection. [A] shows the test error decreases and AUROC increases. [B] also shows test error decreases, while AUROC decreases dramatically. [C] shows the test accuracy is stable while AUROC shows instability. We avoid these two drawbacks by AoP, visualized in (b). These claims are confirmed by experiments shown in (c)-(d). The model is a ResNet-18~\cite{he_deep_2016} trained on CIFAR-10~\cite{krizhevsky2009learning}, and OOD dataset is LSUN~\cite{yu_2015_lsun}.}	
	\label{fig:ood_overfitting}
	\vskip -0.15in
\end{figure}

One core issue of detecting OOD samples is to design a detector $\phi \left( \boldsymbol{x} \right)$ that maps an input $\boldsymbol{x}$ to a scalar to distinguish OOD data from in-distribution (ID). In the test stage, given a threshold $\delta$, samples are regarded as OOD data if $\phi \left( \boldsymbol{x} \right) >\delta$, and as ID otherwise. By changing the threshold, AUROC, i.e., the area under Receiver Operating Characteristics (ROC) curve is calculated, which is a threshold-independent metric in OOD detection.

Recently, there has been great progress in detecting OOD data. Various scoring functions are
designed to distinctly distinguish the output scores of ID and OOD data, like ODIN~\cite{liang_2018_ODIN}, Mahalanobis distance~\cite{lee_2018_maha}, energy score~\cite{liu_2020_energy} and ViM~\cite{wang2022vim}. Besides, one of the most effective methods up to now, called Outlier Exposure (OE)~\cite{hendrycks_2019_oe}, is to train the models against auxiliary data of natural outlier images. With the rapid advances in model architecture like vision transformer~\cite{dosovitskiy_2021_ViT}, it has also been empirically verified that transformer outperforms convolutional neural network in OOD detection~\cite{fort2021_explorOOD}.

Despite the success of previous methods, less attention has been given to \textit{the chaotic behavior of OOD detection along the optimization trajectory during training}. For example, it remains elusive how the curve of AUROC changes during training. In this paper, we first delve into analyzing the curve of metrics in OOD detection in training stage. As shown in Figure~\ref{fig:ood_overfitting}(a), \textit{\textbf{our findings reveal that the curve of AUROC could be decomposed into three stages}}: In the first stage, i.e. stage [A], both the training error and test error are decreasing, and AUROC is increasing. In the second stage, i.e. stage [B], test error continues decreasing when the training error approaches near zero, while AUROC is decreasing like overfitting. This phenomenon of \textbf{overfitting} greatly harms OOD detection. In the third stage, i.e. stage [C], AUROC varies greatly during training, while the test error is much more stable. The \textbf{instability} of AUROC makes the choice of final model not reliable, because a nearby checkpoint may achieve a much better or worse performance in OOD detection. The overfitting and instability of AUROC during training greatly harm the performance of OOD detection, which is neglected by previous methods.

In this paper, we first analyze the problem of instability. The instability mainly comes from the vulnerability of model to the small perturbations of weight~\cite{chaudhari2017entropysgd} during training. We show that a small perturbation of the final checkpoint could cause AUROC to vary sharply. To alleviate the instability of OOD detection during training, we utilize the moving averaged model as our final model, whose parameters are determined by the moving average of the online model's parameters. Model averaging could lead to a flat loss landscape~\cite{izmailov2018SWA}, which makes AUROC more stable during training. With the help of model averaging, it is more reliable to choose the final checkpoint as the model deployed in practical scenarios.

Next, we investigate why the overfitting happens, and then propose its solution. For an over-parameterized model, continuing training model after reaching near zero training error would drive the model to memorize training data~\cite{zhang2017understanding}. The memorization drives model to learn redundant and noisy features which commonly exist in different datasets, thus causing the overlap of distribution between ID and OOD data. Consequently, the model becomes overfitting in OOD detection. To alleviate the overfitting, our key idea is to employ a global sparsity constraint as a method to evade memorizing training data and learning noisy features, which is closely related to network pruning~\cite{han_2015_pruning}. We theoretically show that LASSO~\cite{tibshirani_1996_lasso} largely eliminates the problem of overfitting based on a linear model, which motivates us to use pruning in neural networks.

Combining the two issues above, we summarize our method as\textit{ Average of Pruning} (\textbf{AoP}), and AoP could be regarded as a simple module readily pluggable into any existing method in OOD detection. It is empirically verified that AoP achieves significant improvement on OOD detection. Our contributions can be summarized as follows:
\begin{itemize}
	\item [$\bullet$] We uncover the phenomenon of instability and overfitting in OOD detection during training, which seriously affects performance. 
	\item [$\bullet$] We verify that instability comes from the vulnerability of model to small changes in weights, and overfitting is reduced by overparameterization which drives model to learn noisy and abundant features.
	\item [$\bullet$] We propose \textit{Average of Pruning}. Model averaging improves stability and pruning eliminates overfitting.
	\item [$\bullet$] We conduct comprehensive experiments and ablation studies to verify the effectiveness of AoP, across different datasets and network architectures.
\end{itemize}

\section{Related Work}
\noindent\textbf{OOD Detection} Various methods are proposed under the setting that whether use the auxiliary data of natural outliers or not. The work of~\cite{hendrycks_2017_baseline} first introduced a baseline called MSP as the scoring function to detect OOD inputs. Later, various methods focus on designing scoring functions are proposed~\cite{liang_2018_ODIN,lee2018ganOutlier,liu_2020_energy,sun_2021_react,wang2022vim,sun2022KNN}. Generative models~\cite{ren2019likeliRatio} are also applied to provide reliable estimations of confidence. Besides, OE~\cite{hendrycks_2019_oe} first introduced an extra dataset consisting of outliers in training and enforced low confidence on such outliers. To better utilize the auxiliary dataset, different methods are propose~\cite{Yang2022Openood,Ming2022poem}. There are also some methods to synthesize outliers~\cite{du2022vos}. 

\noindent\textbf{Model Averaging} The basic idea of model averaging could refer to tail-averaging~\cite{jain2018parallelizing}, which averages the final checkpoints of SGD, and it could decrease the variance induced by SGD noise and stabilize training~\cite{neu2018iterate}. Further, stochastic weight averaging~\cite{izmailov2018SWA} applies cyclical learning rate to average the selected checkpoints. It has been empirically verified that model averaging contributes to a flatter loss landscape and wider minimum~\cite{izmailov2018SWA}. Based on this, we investigate whether OOD detection suffers from instability in training, which is an essential issue neglected before.

\noindent\textbf{Pruning and sparsity} We mainly discuss unstructured pruning. The concept of pruning in neural networks dated back to~\cite{lecun1989optimal}. To reduce the memory in modern neural networks, pruning has attracted great attention~\cite{han_2015_pruning} recently. Methods are proposed to drop the unimportant weights~\cite{han_2015_pruning,zhu2017GMP,evci2020rigl,liu2021GraNet}. The goal of pruning is to deploy modern networks with a slight loss of generalization. Among these, the lottery ticket hypothesis (LTH)~\cite{frankle_2018_LTH,frankle_2020_wrewind} shows that sparse networks training from scratch can reach full accuracy than dense networks. Previous works focus on the generalization of pruned models, while we discuss the connection between sparsity and OOD detection.

\section{Model Averaging Improves Stability}

In this section, we first instability analysis of the final checkpoint on OOD detection. Then, we show that model averaging could largely make the performance more stable.

\subsection{Motivation: Instability Analysis}

\begin{figure}[h]
	\vskip -0.15in
	\begin{center}
		\includegraphics[width=0.99\linewidth]{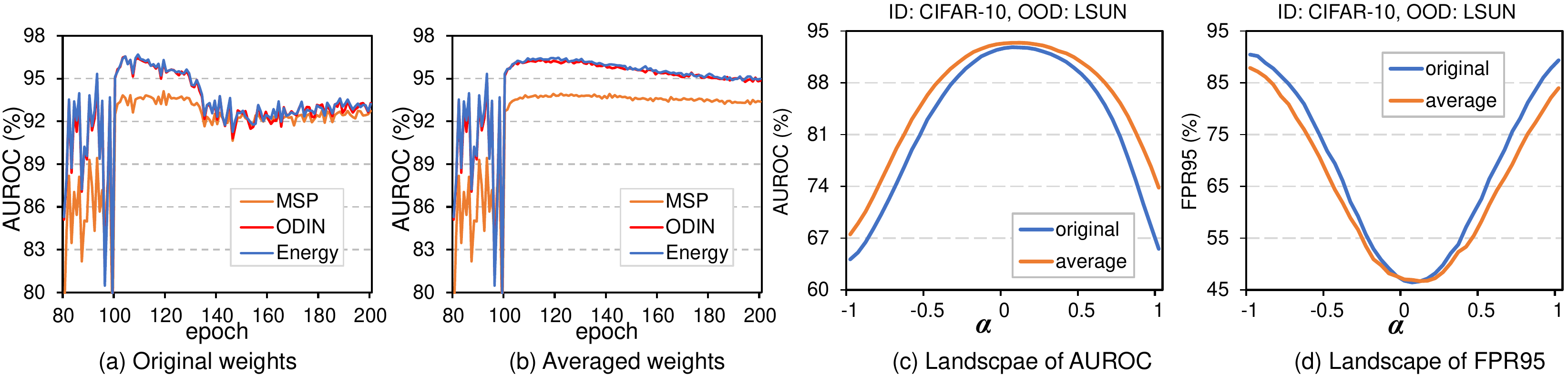}
	\end{center}
	\vskip -0.15in
	\caption{(a)-(b): Influence of different scoring functions. Different scoring function~\cite{liang_2018_ODIN,liu_2020_energy} also suffer from instability. (c)-(d): Landscape of AUROC ($\uparrow$) and FPR95 ($\downarrow$) under weight perturbations for the original model and averaged model.}	
	\label{fig:overfitting_scoring}
	\vskip -0.15in
\end{figure}

Previous methods like scoring functions focus on post-hoc process, while the instability of final checkpoint on OOD detection is neglected. To be more specific, a nearby checkpoint may achieve much better or worse performance. As shown in Fig.~\ref{fig:ood_overfitting}(c), AUROC varies sharply in the final stage, which makes the performance of final checkpoint unreliable. We also find that post-hoc process could not alleviate such instability, as shown in Fig~\ref{fig:overfitting_scoring}(a).

\noindent\textbf{Visualization.} To better investigate how the changes in weights influence the performance of OOD detection, we visualize AUROC and FPR95 landscape by plotting their value changes when moving the weights along a random direction $d$ with magnitude $\alpha$:
$$
\theta\left( d ,t \right) =\theta_{\mathrm{final}}+\alpha \cdot d,
$$
where $d$ is a random direction drawn from a Gaussian distribution and repeated 10 times. The random direction is normalized as proposed in~\cite{li2018visualizing}. The visualization of a trained ResNet-18~\cite{he_deep_2016} is shown in Fig.~\ref{fig:overfitting_scoring}(c).

Based on the landscape of AUROC and FPR95, we find that the solution is fragile under small perturbations of weights, resulting narrow optima in OOD detection. Intuitively, wider optima would make the performance of OOD detection more stable, since the changes of weights may have less influence on performance. 

\subsection{Method: Model Averaging}

To contribute to a wider optimum, we adopt Stochastic Weight Averaging (SWA)~\cite{izmailov2018SWA}, which aims to find much broader optima for better generalization. SWA averages multiple points along the trajectory of SGD with a cyclical or constant learning rate. In this paper, we implement SWA using a simply exponential moving average $\theta_{t}^{\mathrm{MA}}$ of the original model parameters $\theta_{t} $ with a decay rate $\tau$ after epoch $t_0$, called as \textit{Model Averaging} (MA):
\begin{equation}\label{equ:method_MA}
	\begin{aligned}
		\theta _{t+1}^{\mathrm{MA}}=\left\{ \begin{array}{c}
			\theta _t, \,\,\,\, t<t_0       \\
			\tau \cdot \theta _{t}^{\mathrm{MA}}+\left( 1-\tau \right) \cdot \theta _{t+1}, \,\,\,\, t\ge t_0\\
		\end{array} \right .,
	\end{aligned}
\end{equation}
where $\tau =\frac{t-t_0}{t-t_0+1}$ is a constant. In the stage of evaluation, the weighted parameters $\theta _{t}^{\mathrm{MA}}$ are used instead of the originally trained models $\theta_t$. As shown in Fig.~\ref{fig:overfitting_scoring}(b), the averaged model achieves a more stable performance. MA is also shown to stabilize the performance of various scoring functions, as shown in Fig.~\ref{fig:overfitting_scoring}(d). The performance on OOD detection is expected to be smoother since the weights of MA are the average of nearby checkpoints.

MA could be interpreted as approximating FGE ensemble~\cite{garipov2018FGE} with a single model. Ensembling has been verified to boost the performance of OOD detection before~\cite{lakshminarayanan2017ensemble}. However, such ensemble methods require the practicer to train multiple classifiers, which induces much extra computational budget, while MA is more convenient and has almost no computational overhead.

\section{Pruning Eliminates Overfitting}

In this section, we first reveal that an over-parameterized model may cause overfitting in OOD detection during training, and then show that increasing the sparsity of model by pruning could eliminate it based on theoretical analysis.

\subsection{Motivation: Overfitting Analysis}

The training trajectory of models in OOD detection has not been explored clearly before. As shown in Fig.~\ref{fig:ood_overfitting} (a)(c), OOD detection suffers from overfitting, and so does the averaged model in Fig.~\ref{fig:overfitting_scoring}(b). We would like to provide more evidence on such overfitting.

\begin{figure}[h]
	\begin{center}
		\includegraphics[width=0.67\linewidth]{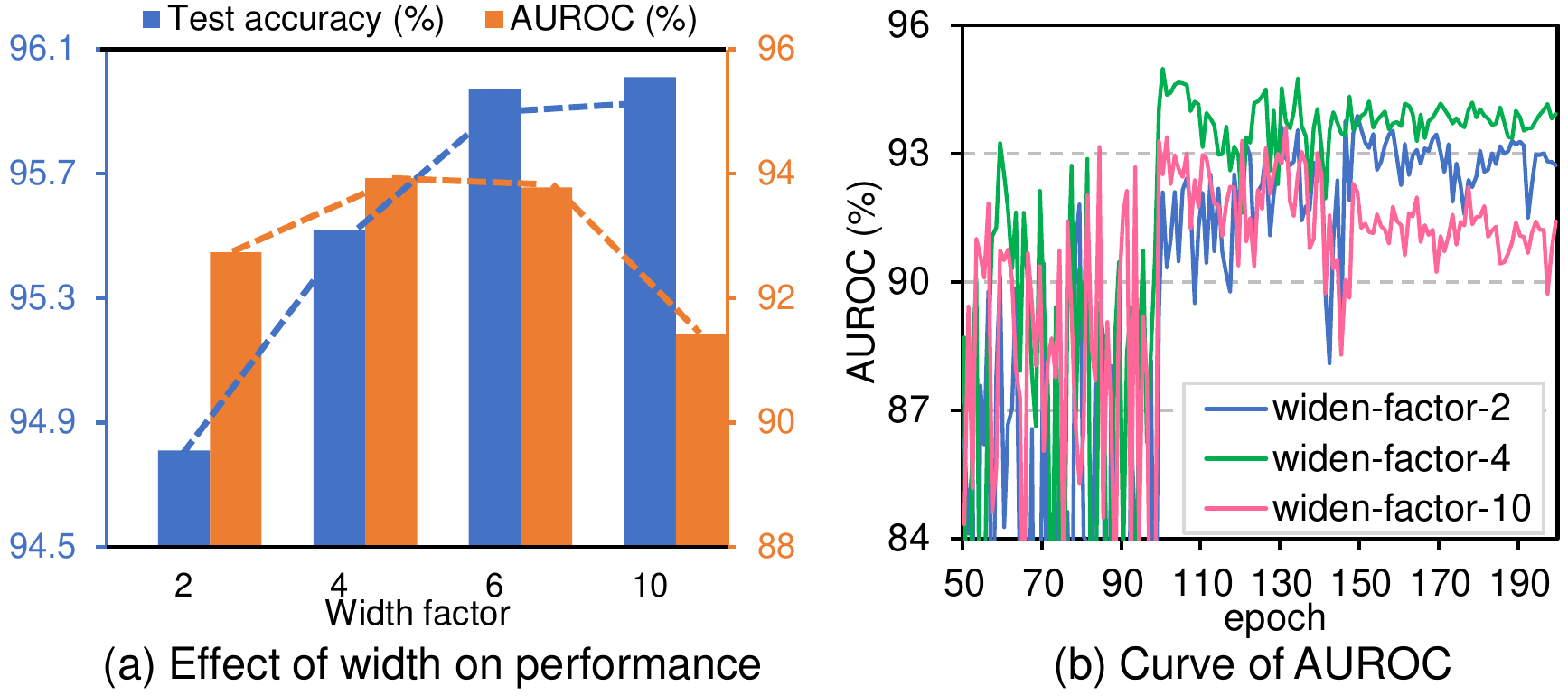}
	\end{center}
	\vskip -0.15in
	\caption{Influence of different width in WideResNet~\cite{zagoruyko_wrn_2017}. (a) As width increases, test accuracy increases while AUROC first increases and then decreases. (b) Curves of AUROC during training.}	
	\label{fig:overfitting_diemension}
	\vskip -0.05in
\end{figure}

\setParDis\noindent\textbf{More empirical verifications.} We conduct experiments on WideResNet-28~\cite{zagoruyko_wrn_2017} with various width factors, and the results are shown in Fig~\ref{fig:overfitting_diemension}. We can see as the width increases, the test accuracy increases steadily while AUROC first increases and then decreases. This verifies that OOD detection is harmed by overfitting from overparameterization.

\setParDis\noindent\textbf{Theoretical Analysis.} Suppose we have an ID dataset $\mathcal{D} _{\mathrm{ID}}$ consist of samples $(\boldsymbol{x},y)$ and an OOD dataset $\mathcal{D} _{\mathrm{OOD}}$ with samples  $(\boldsymbol{x},q)$.
Formally, we set the $d+1$ dimension features $\boldsymbol{x}=\left( x_1,x_2,\cdots ,x_{d+1} \right)$ of ID and OOD data are drawn according to Gaussian distributions:
\begin{equation}\label{equ:data_distribution}
	\begin{aligned}	
		y&\overset{u.a.r}{\sim}\left\{ -1,+1 \right\} ,\,\,\, \boldsymbol{x}\sim \mathcal{N} \left( y\cdot \boldsymbol{\mu } _{\mathrm{ID}},\sigma ^2 \boldsymbol{I} \right) 
		\\
		q&\overset{u.a.r}{\sim}\left\{ -1,+1 \right\} ,\,\,\, \boldsymbol{x}\sim \mathcal{N} \left( q\cdot \boldsymbol{\mu } _{\mathrm{OOD}},\sigma ^2 \boldsymbol{I} \right)
	\end{aligned},	
\end{equation}
where $\boldsymbol{\mu }_{\mathrm{ID}}=\left( 1,\eta ,\cdots ,\eta \right)$ and $\boldsymbol{\mu }_{\mathrm{OOD}}=\left( 0,\eta ,\cdots ,\eta \right) \in \mathbb{R} ^{d+1}$. The distribution of $(x_1)$ is different for ID and OOD data, which is called as \textit{special feature} because it reveals the difference between ID and OOD data. The distribution of the rest features $ \left( x_2,\dots ,x_{d+1} \right) $ are the same for ID and OOD data, which are called as \textit{common features} because they may induce overlap for ID and OOD data. Usually, the common features mean various noise rather than semantic information, so they are often less discriminative but abundant~\cite{ilyas_adversarial_2019,tao2021better,tao2022can}, i.e., $0<\eta \ll 1$ and $d\gg 1$.

The Bayes optimal classifier $f_{\mathrm{Bayes}}\left( \boldsymbol{x} \right)$ could be derived based on the distribution of ID data. Then, the decision is based on the sign of logits from Bayes classifier, i.e.,  
\begin{equation}\label{equ:bayes_classifier}
	f_{\mathrm{Bayes}}\left( \boldsymbol{x} \right) =x_1+\eta \sum_{i=2}^{d+1}{x_i}.
\end{equation}

\begin{definition}[ID classification risk, OOD detection risk]\label{def:alignment}
	For a classifier $f$, the risks induced by classification on ID data $\mathcal{R} _{\mathrm{ID}}\left( f \right)$ and rejection on OOD data $\mathcal{R} _{\mathrm{OOD}}\left( f\right)$ are
	\begin{equation}\label{equ:risk}
		\begin{aligned}
			&	\mathcal{R} _{\mathrm{ID}}\left( f \right) =\underset{\left( \boldsymbol{x},y \right) \sim \mathcal{D} _{\mathrm{ID}}}{\mathrm{Pr}}\left\{ \mathrm{sign}(f\left( \boldsymbol{x} \right)) \ne y \right\}, 
			\\
			&\mathcal{R} _{\mathrm{OOD}}\left( f \right) =\underset{\left( \boldsymbol{x},q \right) \sim \mathcal{D} _{\mathrm{OOD}}}{\mathrm{Pr}}\left\{ \left| f\left( \boldsymbol{x} \right) \right|>\delta \right\},
		\end{aligned}
	\end{equation}
	where $\delta$ is a given threshold for detecting OOD data. Inputs are accepted as ID data when the magnitude of their logits are larger than the threshold. 
\end{definition}
\begin{theorem}\label{pro:simplified_risk}
	Suppose the distributions of ID and OOD data follow Gaussian distribution in Eq.~(\ref{equ:data_distribution}), then $\mathcal{R} _{\mathrm{ID}}\left( f_{\mathrm{Bayes}} \right)$ and $\mathcal{R} _{\mathrm{OOD}}\left( f_{\mathrm{Bayes}} \right)$ could be simplified to
	\begin{equation}\label{equ:risk_simplified}
		\begin{aligned}
			&\mathcal{R} _{\mathrm{ID}}\left( f_{\mathrm{Bayes}} \right) =\,\,\mathrm{Pr}\left\{ \mathcal{N} \left( 0,1 \right) >\frac{\sqrt{1+d\eta ^2}}{\sigma} \right\},
			\\
			&\mathcal{R} _{\mathrm{OOD}}\left( f_{\mathrm{Bayes}} \right) =\mathrm{Pr}\left\{ \mathcal{N} \left( 0,1 \right) >\frac{\delta -d\eta ^2}{\sigma \cdot \sqrt{1+d\eta ^2}} \right\} +\mathrm{Pr}\left\{ \mathcal{N} \left( 0,1 \right) >\frac{\delta +d\eta ^2}{\sigma \cdot \sqrt{1+d\eta ^2}} \right\}.
		\end{aligned}
	\end{equation}
\end{theorem}

\begin{figure}[h]
	\vskip -0.10in
	\begin{center}
		\includegraphics[width=1.00\linewidth]{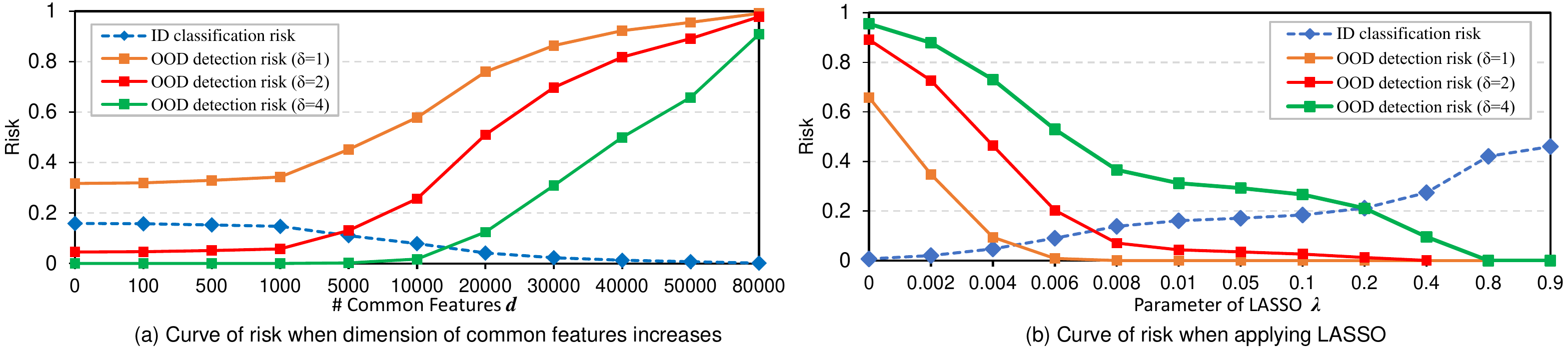}
	\end{center}
	\vskip -0.15in
	\caption{A detailed example on how the number of common features $d$ and LASSO influence $\mathcal{R} _{\mathrm{ID}}\left( f \right)$ (i.e., ID classification risk) and $\mathcal{R} _{\mathrm{OOD}}\left( f \right)$ (i.e., OOD detection risk). As a concrete example, here we set $\sigma=1,\eta=0.01$. In (b), we keep $d=50000$.}	
	\label{fig:theoretical_results}
	\vskip -0.05in
\end{figure}

Theorem~\ref{pro:simplified_risk} indicates that the number of common features $d$ has a close connection with accuracy and OOD detection performance. For clearer understanding, a detailed example is shown in Fig~\ref{fig:theoretical_results}(a) with concrete settings. As $d$ increases, $ \mathcal{R} _{\mathrm{ID}}\left( f_{\mathrm{Bayes}} \right) $ decreases slowly while $\mathcal{R} _{\mathrm{OOD}}\left( f_{\mathrm{Bayes}} \right)$ increases sharply. We could conclude that an approximately small value of $d$ contributes a better trade-off between classification and OOD detection. The curves of OOD detection risk with different rejection thresholds $\delta$ are also given, and they show a consistent tendency.

If the model is over-parameterized, it is reasonable to use feature selection to reduce the over-much common features to improve performance on OOD detection. We would show that the well-known method in feature selection LASSO~\cite{tibshirani_1996_lasso} could achieve this goal. The solution with LASSO could be obtain by optimizing the overall loss~\cite{hastie2009esl}:
\begin{equation}\label{equ:lasso}
	\underset{\boldsymbol{w}}{\min}\,\,\mathbb{E} _{\left( \boldsymbol{x},y \right) \sim \mathcal{D} _{\mathrm{ID}}}\left\{ \frac{1}{2} \left( \boldsymbol{w}^T\frac{\boldsymbol{x}}{\sigma}-y \right) ^2+\lambda \cdot \left| \boldsymbol{w} \right|_1 \right\},
\end{equation}
where $\lambda$ is the parameter of LASSO.
\begin{theorem}\label{pro:LASSO_solution}
	The classifier $f_{\mathrm{LASSO}}$~\cite{hastie2009esl} obtained by minimizing the loss in Eq.~(\ref{equ:lasso}) is
	\begin{equation}\label{equ:lasso_classifier}
		f_{\mathrm{LASSO}}\left( \boldsymbol{x} \right) = \left( \left( 1-\lambda \right) _+\cdot x_1+\sum_{i=2}^{d+1}{\left( \eta -\lambda \right) _+\cdot x_i} \right),
	\end{equation}
	where $\left( \cdot \right) _+$ means the positive part of a number.
\end{theorem}

For intuitive understanding of Theorem~\ref{pro:LASSO_solution}, we draw $\mathcal{R} _{\mathrm{ID}}$ and $\mathcal{R} _{\mathrm{OOD}}$ for $f_{\mathrm{LASSO}}$ while changing the parameter $\lambda$ in LASSO, and the results are shown in Fig.~\ref{fig:theoretical_results}(b). As $\lambda$ increases, $ \mathcal{R} _{\mathrm{OOD}} $ rapidly decreases and $ \mathcal{R} _{\mathrm{ID}} $ increases. A better trade-off could be achieved while $\lambda \approx \eta =0.01$.

\subsection{Method: Lottery Tickets Hypothesis} 

As shown in the previous analysis, overparameterization causes the models to learn more noisy and redundant features, which results in overfitting in OOD detection. We also show LASSO~\cite{tibshirani_1996_lasso} is a promising method. However, LASSO is not suitable for neural networks. For deep models, pruning aims to construct sparse models.

For a neural network $f\left( x;\theta \right)$ with parameters $\theta$, its pruned model could be denoted as $f\left( x;m\odot \theta \right)$ with a set of binary masks $m\in \left\{ 0,1 \right\} ^n$, where $\odot$ means the element-wise product and $n$ is the number of parameters. One effective method is the Lottery Ticket Hypothesis (LTH)~\cite{frankle_2018_LTH}, which adopts iterative magnitude pruning (IMP).

In this paper, we adopt the commonly used LTH with weight rewinding~\cite{frankle_2020_wrewind} to find the sparse subnetworks. Pseudo-code is provided in Algorithm~\ref{alg:IMP}, and it seems similar to the solution of LASSO for the linear model in Eq.~(\ref{equ:lasso_classifier}). 

\begin{figure}[h]
	\vskip -0.10in
	\begin{center}
		\includegraphics[width=0.70\linewidth]{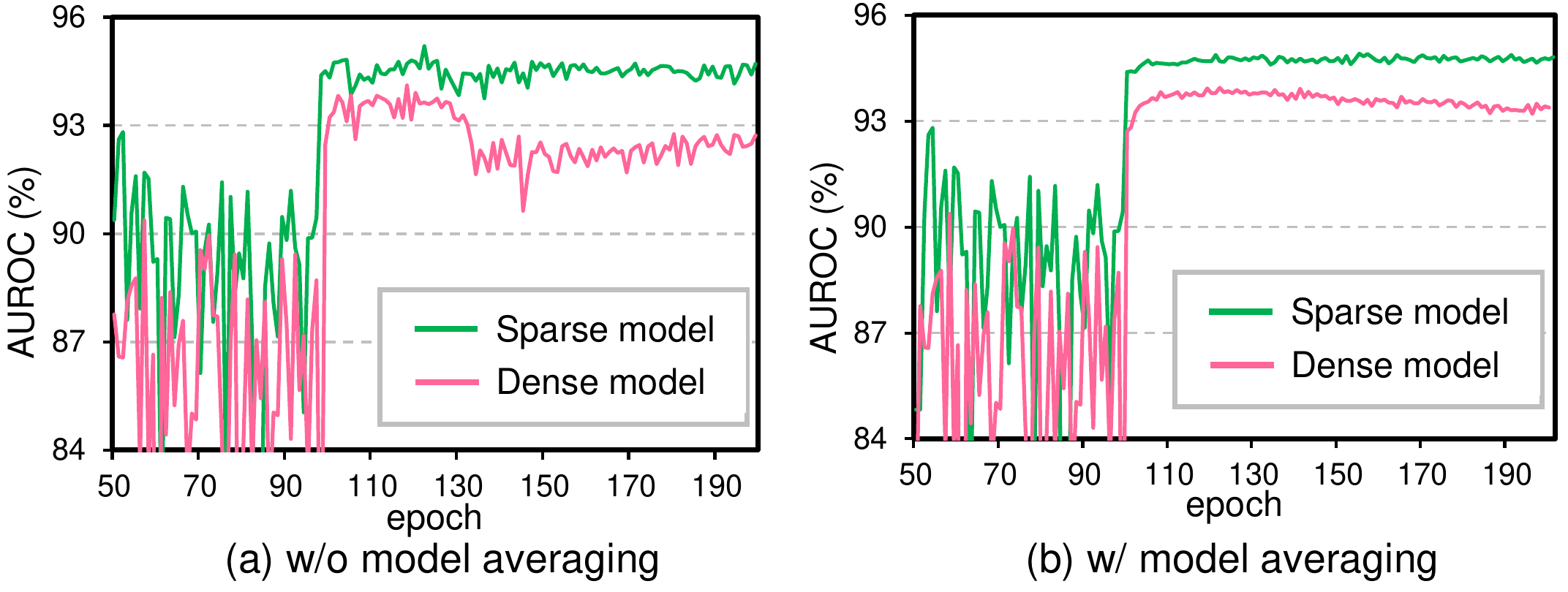}
	\end{center}
	\vskip -0.18in
	\caption{Effect of pruning on eliminating overfitting. The settings are the same as Fig.~\ref{fig:ood_overfitting}.}	
	\label{fig:overfitting_pruning}
	\vskip -0.07in
\end{figure}

In the original paper of LTH~\cite{frankle_2018_LTH}, it is claimed that dense randomly-initialized neural networks contain a sparse subnetwork that can be trained in isolation to match the test accuracy of the original network. However, this hypothesis has not been introduced to OOD detection. In this paper, we propose that pruning~\cite{han_2015_pruning} (e.g, LTH) also helps OOD detection by alleviating overfitting. As shown in Fig~\ref{fig:overfitting_pruning}, when we increase the sparsity of model, the overfitting in OOD detection during training becomes weaker, and thus we achieve a better performance in the final checkpoint.

In practical applications, large networks are often compressed to reduce the number of weights or the memory footprint~\cite{han_2015_pruning} because of the memory limits. Our results show that one of the byproducts of a pruned model is the improvement in detecting unknown inputs,  which is essential in practical scenarios. We will also provide comparisons with other pruning methods in the latter section. 

\begin{algorithm}[t]
	\caption{IMP weight rewinding for LTH}
	\label{alg:IMP}
	\renewcommand{\algorithmicrequire}{\textbf{Input:}}
	\renewcommand{\algorithmicensure}{\textbf{Output:}}
	\begin{algorithmic}
		\REQUIRE Rewinding step $k$, training steps $T$, pruning iterations $N$.
		\ENSURE Learned weights $\theta$, pruning mask $m$.
		\STATE Randomly initialize network $f$ with initial weights $\theta_0\in\mathbb{R}^d$.
		\STATE Initialize pruning mask to $m=1^d$.
		\STATE Train $\theta_0$ to $\theta_k$ for $k$ steps.
		\FOR {$n=1,\cdots ,N$}
		\STATE Train $m\odot \theta_k$ to $m\odot \theta_T$.
		\STATE Prune the 20\% lowest-magnitude weights globally and get mask $m$.
		\STATE Rewind the weights to $\theta_k$.
		\ENDFOR
	\end{algorithmic}
\end{algorithm}

\subsection{Overall Method: Average of Pruning}
\label{sec:Winning Tickets in OOD Detection}

Combining model averaging and pruning, we propose \textit{Average of Pruning} (dubbed \textbf{AoP}) to boost OOD detection performance. As many of the effective OOD detection methods are post-hoc processes, AoP could be easily applied as a simple module readily pluggable into any existing method. The pipeline is divided as follows:
\begin{tcolorbox}[colback=gray!10]
	\textbf{Step 1}: Given an initialized network, we train a sparse network by LTH, according to Algorithm~\ref{alg:IMP}.	
	
	\textbf{Step 2}: While reaching the desired sparsity, we calculate the averaged model by Equation~\ref{equ:method_MA}. 
	
	\textbf{Step 3}: Various post-hoc methods could be applied to the averaged model. 	
\end{tcolorbox}

\section{Experiments}

In this section, we evaluate AoP on various OOD detection tasks. We first verify the effectiveness of AoP on CIFAR benchmarks and provide comprehensive ablation studies. Then, we proceed with ImageNet benchmark.

\subsection{Evaluation on CIFAR Benchmarks}
\label{sec:cifar_benchmarks}
\noindent\textbf{In-distribution Datasets.} We use CIFAR-10 and CIFAR-100~\cite{krizhevsky2009learning} as in-distribution data.

\begin{table*}[t]
	\centering
	\caption{Performance of OOD detection for various methods on CIFAR benchmarks. All the metrics on OOD detection are the \textit{average on six OOD datasets}. All values are percentages. The best results are boldfaced for highlight.}
	\vskip -0.10in
	\label{tab:main_results}
	\renewcommand{\arraystretch}{1.3}
	\renewcommand\tabcolsep{5.0pt}
	\scalebox{0.60}{
		\begin{tabular}{ccccccccccc}
			\toprule[1.4pt]
			\multirow{2}{*}{\textbf{Model}} &  \multirow{2}{*}{\textbf{Methods}} & \multirow{2}{*}{\textbf{Reference}}   & \multicolumn{4}{c}{In-distribution dataset: \textbf{CIFAR-10}}    & \multicolumn{4}{c}{In-distribution dataset: \textbf{CIFAR-100}}   \\ \cmidrule(r){4-7}  \cmidrule(r){8-11}
			&    &    & \multicolumn{1}{c}{\textbf{AUROC} $\uparrow$} & \multicolumn{1}{c}{\textbf{AUPR} $\uparrow$} & \multicolumn{1}{c}{\textbf{FPR95} $\downarrow$} & \textbf{Acc} $\uparrow$ & \multicolumn{1}{c}{\textbf{AUROC} $\uparrow$} & \multicolumn{1}{c}{\textbf{AUPR} $\uparrow$} & \multicolumn{1}{c}{\textbf{FPR95} $\downarrow$} & \textbf{Acc} $\uparrow$                                    
			\\  \hline
			\multicolumn{1}{c}{\multirow{11}{*}{\textbf{ResNet-18}}} & \multicolumn{1}{l}{MSP~\cite{hendrycks_2017_baseline}}  & \multicolumn{1}{l}{ICLR2017}
			& \multicolumn{1}{c}{91.63 $\pm$ 0.26}      & \multicolumn{1}{c}{98.08 $\pm$ 0.11}     & \multicolumn{1}{c}{50.00 $\pm$ 0.50}      &  94.98 $\pm$ 0.11   & \multicolumn{1}{c}{76.64 $\pm$ 0.20}      & \multicolumn{1}{c}{94.31 $\pm$ 0.08}     & \multicolumn{1}{c}{80.03 $\pm$ 0.72}      &  75.83 $\pm$ 0.33                      
			\\ 
			& \multicolumn{1}{l}{ODIN~\cite{liang_2018_ODIN}}  & \multicolumn{1}{l}{ICLR2018}
			& \multicolumn{1}{c}{92.13 $\pm$ 0.96}      & \multicolumn{1}{c}{97.92 $\pm$ 0.36}     & \multicolumn{1}{c}{36.19 $\pm$ 3.47}  &  94.98 $\pm$ 0.11  & \multicolumn{1}{c}{81.05 $\pm$ 0.83}      & \multicolumn{1}{c}{95.44 $\pm$ 0.18}     & \multicolumn{1}{c}{74.04 $\pm$ 2.53}   &   75.83 $\pm$ 0.33                         
			\\ 
			& \multicolumn{1}{l}{Maha~\cite{lee_2018_maha}}  & \multicolumn{1}{l}{NeurIPS2018}
			& \multicolumn{1}{c}{90.61 $\pm$ 1.13}      & \multicolumn{1}{c}{97.88 $\pm$ 0.24}     & \multicolumn{1}{c}{45.31 $\pm$ 4.63}  & 94.98 $\pm$ 0.11  & \multicolumn{1}{c}{74.36 $\pm$ 0.64}      & \multicolumn{1}{c}{92.88 $\pm$ 0.13}     & \multicolumn{1}{c}{ 71.97$\pm$ 2.15}  &  75.83 $\pm$ 0.33                          
			\\ 	
			& \multicolumn{1}{l}{Energy~\cite{liu_2020_energy}}  &  \multicolumn{1}{l}{NeurIPS2020}
			& \multicolumn{1}{c}{92.24 $\pm$ 1.02}      & \multicolumn{1}{c}{97.96 $\pm$ 0.35}     & \multicolumn{1}{c}{35.15 $\pm$ 3.43}      & 94.98 $\pm$ 0.11    & \multicolumn{1}{c}{81.36 $\pm$ 0.87}      & \multicolumn{1}{c}{95.50 $\pm$ 0.18}     & \multicolumn{1}{c}{72.29 $\pm$ 2.78}   &  75.83 $\pm$ 0.33                         
			\\ 
			& \multicolumn{1}{l}{GradNorm~\cite{huang2021gradnorm}}  & \multicolumn{1}{l}{NeurIPS2021}
			& \multicolumn{1}{c}{50.04 $\pm$ 5.55}      & \multicolumn{1}{c}{81.79 $\pm$ 2.59}     & \multicolumn{1}{c}{84.18 $\pm$ 2.53}   &  94.98 $\pm$ 0.11   & \multicolumn{1}{c}{69.78 $\pm$ 3.63}    & \multicolumn{1}{c}{90.69 $\pm$ 1.81}     & \multicolumn{1}{c}{74.91 $\pm$ 1.08}   &  75.83 $\pm$ 0.33               
			\\
			& \multicolumn{1}{l}{ReAct~\cite{sun_2021_react}}  & \multicolumn{1}{l}{NeurIPS2021}
			& \multicolumn{1}{c}{92.46 $\pm$ 0.73}      & \multicolumn{1}{c}{98.19 $\pm$ 0.52}     & \multicolumn{1}{c}{34.63 $\pm$ 2.80}      &  94.98 $\pm$ 0.11   & \multicolumn{1}{c}{80.35 $\pm$ 0.52}      & \multicolumn{1}{c}{95.36 $\pm$ 0.11}   & \multicolumn{1}{c}{76.20 $\pm$ 2.01}   &   75.83 $\pm$ 0.33                            
			\\ 
			& \multicolumn{1}{l}{MaxLogit~\cite{hendrycks2022MaxL}}  & \multicolumn{1}{l}{ICML2022}
			& \multicolumn{1}{c}{92.61 $\pm$ 0.29}      & \multicolumn{1}{c}{98.13 $\pm$ 0.07}     & \multicolumn{1}{c}{35.09 $\pm$ 1.90}  &  94.98 $\pm$ 0.11   & \multicolumn{1}{c}{81.02 $\pm$ 0.84}      & \multicolumn{1}{c}{95.44 $\pm$ 0.17}   & \multicolumn{1}{c}{74.13 $\pm$ 2.64}   &   75.83 $\pm$ 0.33                           
			\\ 	
			& \multicolumn{1}{l}{VOS~\cite{wei2022logitnorm}}    & \multicolumn{1}{l}{ICLR2022}
			& \multicolumn{1}{c}{93.54 $\pm$ 1.05}      & \multicolumn{1}{c}{98.47 $\pm$ 0.28}     & \multicolumn{1}{c}{32.90 $\pm$ 5.02} &  94.67 $\pm$ 0.14 & \multicolumn{1}{c}{81.14 $\pm$ 1.02}      & \multicolumn{1}{c}{95.40 $\pm$ 0.36}     & \multicolumn{1}{c}{74.46 $\pm$ 3.54}      & 74.66 $\pm$  0.26        
			\\ 	 \cmidrule(r){2-3}   \cmidrule(r){4-7}  \cmidrule(r){8-11}
			
			& \multicolumn{1}{l}{KNN~\cite{sun2022KNN}}    & \multicolumn{1}{l}{ICML2022}
			& 94.39 $\pm$ 0.27  & 98.65 $\pm$ 0.24   & 34.42 $\pm$ 0.58      & \textbf{94.98 $\pm$ 0.11}  & 80.43 $\pm$ 0.88  & 94.84 $\pm$ 0.32    & 69.72 $\pm$ 1.84   &  75.83 $\pm$ 0.33                          
			\\ 	
			& \multicolumn{1}{l}{$\quad$ + \textbf{AoP}}   &  -
			& \cellcolor{gray!30} \textbf{95.04 $\pm$ 0.17}     & \cellcolor{gray!30} \textbf{98.88 $\pm$ 0.04}     & \cellcolor{gray!30} \textbf{25.30 $\pm$ 1.63} \footnotesize {\textcolor{magenta}{$\downarrow$ \textbf{9.1}}} &  \cellcolor{gray!30} 94.89 $\pm$ 0.11  & \cellcolor{gray!30} \textbf{84.36 $\pm$ 0.71}     & \cellcolor{gray!30} \textbf{96.12 $\pm$ 0.22}     & \cellcolor{gray!30} \textbf{58.01 $\pm$ 1.08}  \footnotesize {\textcolor{magenta}{$\downarrow$ \textbf{11.7}}}    &  \cellcolor{gray!30} \textbf{76.08 $\pm$ 0.15}  
			\\ 	 \cmidrule(r){2-3}   \cmidrule(r){4-7}  \cmidrule(r){8-11}
			& \multicolumn{1}{l}{ViM~\cite{wang2022vim}}    & \multicolumn{1}{l}{CVPR2022}
			& 94.18 $\pm$ 0.26      & 98.68 $\pm$ 0.04     & 32.27 $\pm$ 1.02  & \textbf{94.98 $\pm$ 0.11}    & 81.03 $\pm$ 1.84     & 95.53 $\pm$ 0.45    & 72.90 $\pm$ 5.91   &  75.83 $\pm$ 0.33                          
			\\ 	
			& \multicolumn{1}{l}{\textbf{$\quad$ + \textbf{AoP}}}   &  -
			& \cellcolor{gray!30} \textbf{95.69 $\pm$ 0.19}  & \cellcolor{gray!30} \textbf{99.08 $\pm$ 0.04}    & \cellcolor{gray!30} \textbf{25.3 $\pm$ 1.63} \footnotesize {\textcolor{magenta}{$\downarrow$ \textbf{7.0}}} &  \cellcolor{gray!30} 94.89 $\pm$ 0.11  & \cellcolor{gray!30} \textbf{86.16 $\pm$ 0.71}      & \cellcolor{gray!30} \textbf{96.76 $\pm$ 0.18}  & \cellcolor{gray!30} \textbf{59.28 $\pm$ 2.77} \footnotesize {\textcolor{magenta}{$\downarrow$ \textbf{13.6}}} & \cellcolor{gray!30} \textbf{76.08 $\pm$ 0.15}        
			\\ 	\hline
			
			\toprule[1.4pt]	
			\multicolumn{1}{c}{\multirow{11}{*}{\textbf{WRN-28-10}}} & \multicolumn{1}{l}{MSP~\cite{hendrycks_2017_baseline}}   & \multicolumn{1}{l}{ICLR2017}
			& \multicolumn{1}{c}{91.24 $\pm$ 0.08}      & \multicolumn{1}{c}{97.72 $\pm$ 0.05}     & \multicolumn{1}{c}{46.21 $\pm$ 1.72}  & 95.70 $\pm$ 0.07  & \multicolumn{1}{c}{75.77 $\pm$ 0.58}   & \multicolumn{1}{c}{93.92 $\pm$ 0.26}    & \multicolumn{1}{c}{81.53 $\pm$ 0.40}   & 79.72 $\pm$ 0.22         \\ 
			& \multicolumn{1}{l}{ODIN~\cite{liang_2018_ODIN}}  & \multicolumn{1}{l}{ICLR2018}
			& \multicolumn{1}{c}{92.35 $\pm$ 0.31}      & \multicolumn{1}{c}{97.78 $\pm$ 0.13}     & \multicolumn{1}{c}{30.63 $\pm$ 0.86}   &  95.70 $\pm$ 0.07   & \multicolumn{1}{c}{81.25 $\pm$ 0.68}      & \multicolumn{1}{c}{95.42 $\pm$ 0.19}     & \multicolumn{1}{c}{74.28 $\pm$ 1.64}      &   79.72 $\pm$ 0.22     \\ 
			& \multicolumn{1}{l}{Maha~\cite{lee_2018_maha}}  & \multicolumn{1}{l}{NeurIPS2018} 
			& \multicolumn{1}{c}{94.30 $\pm$ 0.37}      & \multicolumn{1}{c}{98.75 $\pm$ 0.08}     & \multicolumn{1}{c}{28.82 $\pm$ 1.94}   &  95.70 $\pm$ 0.07   & \multicolumn{1}{c}{63.32 $\pm$ 1.46}      & \multicolumn{1}{c}{88.83 $\pm$ 0.57}     & \multicolumn{1}{c}{79.36 $\pm$ 1.08}      &  79.72 $\pm$ 0.22   \\ 	
			& \multicolumn{1}{l}{Energy~\cite{liu_2020_energy}}  & \multicolumn{1}{l}{NeurIPS2020}
			& \multicolumn{1}{c}{92.49 $\pm$ 0.32}      & \multicolumn{1}{c}{97.82 $\pm$ 0.14}     & \multicolumn{1}{c}{29.72 $\pm$ 0.99}   &  95.70 $\pm$ 0.07   & \multicolumn{1}{c}{78.03 $\pm$ 0.43}      & \multicolumn{1}{c}{94.47 $\pm$ 0.11}     & \multicolumn{1}{c}{78.65 $\pm$ 1.40}      &  79.72 $\pm$ 0.22      \\ 
			& \multicolumn{1}{l}{GradNorm~\cite{huang2021gradnorm}}  & \multicolumn{1}{l}{NeurIPS2021}
			& \multicolumn{1}{c}{47.70 $\pm$ 2.68}      & \multicolumn{1}{c}{85.35 $\pm$ 5.52}     & \multicolumn{1}{c}{87.45 $\pm$ 2.02}   &  95.70 $\pm$ 0.07   & \multicolumn{1}{c}{64.90 $\pm$ 2.60}      & \multicolumn{1}{c}{89.18 $\pm$ 1.07}     & \multicolumn{1}{c}{81.82 $\pm$ 2.40}      &  79.72 $\pm$ 0.22       \\ 
			& \multicolumn{1}{l}{ReAct~\cite{sun_2021_react}}  & \multicolumn{1}{l}{NeurIPS2021}
			& \multicolumn{1}{c}{82.80 $\pm$ 1.60}      & \multicolumn{1}{c}{95.93 $\pm$ 0.33}     & \multicolumn{1}{c}{69.93 $\pm$ 6.42}   &  95.70 $\pm$ 0.07   & \multicolumn{1}{c}{84.09 $\pm$ 1.58}      & \multicolumn{1}{c}{96.33 $\pm$ 0.38}     & \multicolumn{1}{c}{68.46 $\pm$ 5.91}      &  79.72 $\pm$ 0.22        
			\\
			& \multicolumn{1}{l}{MaxLogit~\cite{hendrycks2022MaxL}} & \multicolumn{1}{l}{ICML2022}
			& \multicolumn{1}{c}{92.35 $\pm$ 0.27}      & \multicolumn{1}{c}{97.78 $\pm$ 0.13}     & \multicolumn{1}{c}{30.48 $\pm$ 0.87}   &  95.70 $\pm$ 0.07   & \multicolumn{1}{c}{77.82 $\pm$ 0.56}      & \multicolumn{1}{c}{94.41 $\pm$ 0.16}     & \multicolumn{1}{c}{79.40 $\pm$ 1.18}      & 79.72 $\pm$ 0.22         
			\\ 
			& \multicolumn{1}{l}{VOS~\cite{du2022vos}} & \multicolumn{1}{l}{ICLR2022}
			& \multicolumn{1}{c}{91.01 $\pm$ 1.49}      & \multicolumn{1}{c}{95.46 $\pm$ 2.42}     & \multicolumn{1}{c}{30.98 $\pm$ 4.65}  &  95.70 $\pm$ 0.07   & \multicolumn{1}{c}{80.84 $\pm$ 1.34}      & \multicolumn{1}{c}{95.24 $\pm$ 0.40}     & \multicolumn{1}{c}{73.66 $\pm$ 3.31}      &  79.79 $\pm$ 0.20          
			\\  \cmidrule(r){2-3}   \cmidrule(r){4-7}  \cmidrule(r){8-11}
			
			& \multicolumn{1}{l}{KNN~\cite{sun2022KNN}}  & \multicolumn{1}{l}{ICML2022}
			& 94.40 $\pm$ 0.14      & 98.79 $\pm$ 0.04     & 32.74 $\pm$ 0.82   &  95.70 $\pm$ 0.07   & 80.50 $\pm$ 0.56    & 95.03 $\pm$ 0.26    & 69.68 $\pm$ 0.91     &  79.72 $\pm$ 0.22                        
			\\ 	
			& \multicolumn{1}{l}{$\quad$ + \textbf{AoP}}  & -
			& \cellcolor{gray!30}\textbf{95.89 $\pm$ 0.03}  & \cellcolor{gray!30}\textbf{99.12 $\pm$ 0.02}     & \cellcolor{gray!30}\textbf{23.95 $\pm$ 0.08}  \footnotesize {\textcolor{magenta}{$\downarrow$ \textbf{8.8}}}    &  \cellcolor{gray!30}\textbf{96.00 $\pm$ 0.15}  & \cellcolor{gray!30}\textbf{83.61 $\pm$ 0.30}    & \cellcolor{gray!30}\textbf{95.88 $\pm$ 0.11}    & \cellcolor{gray!30}\textbf{63.04 $\pm$ 1.06}  \footnotesize {\textcolor{magenta}{$\downarrow$ \textbf{6.6}}}    &   \cellcolor{gray!30}\textbf{80.36 $\pm$ 0.18} 
			\\ 	  \cmidrule(r){2-3}   \cmidrule(r){4-7}  \cmidrule(r){8-11}
			& \multicolumn{1}{l}{ViM~\cite{wang2022vim}}  & \multicolumn{1}{l}{CVPR2022}
			& 96.31 $\pm$ 0.28     & 99.12 $\pm$ 0.08   & 18.35 $\pm$ 1.46  & 95.70 $\pm$ 0.07   & 84.04 $\pm$ 0.95   & 96.13 $\pm$ 0.28   & 62.41 $\pm$ 1.58     &  79.72 $\pm$ 0.22                       
			\\ 	 
			& \multicolumn{1}{l}{$\quad$ + \textbf{AoP}}  & -
			& \cellcolor{gray!30}\textbf{96.53 $\pm$ 0.20}      & \cellcolor{gray!30} \textbf{99.14 $\pm$ 0.06}     & \cellcolor{gray!30}\textbf{15.83 $\pm$ 0.24}  \footnotesize {\textcolor{magenta}{$\downarrow$ \textbf{2.7}}}   &  \cellcolor{gray!30}96.00 $\pm$ 0.15  & \cellcolor{gray!30}\textbf{85.72 $\pm$ 0.45}     & \cellcolor{gray!30}\textbf{96.60 $\pm$ 0.16}     & \cellcolor{gray!30}\textbf{58.71 $\pm$ 1.03}  \footnotesize {\textcolor{magenta}{$\downarrow$ \textbf{3.7}}}   &   \cellcolor{gray!30}\textbf{80.36 $\pm$ 0.18}     
			\\ 
			\toprule[1.4pt]	
	\end{tabular}}
	\vskip -0.15in
\end{table*}

\setParDis\noindent\textbf{Out-of-distribution Datasets.}  For the OOD datasets, we use six common benchmarks as used in previous study~\cite{liu_2020_energy,sun_2021_react,salehi_2021_OODsurvey}: \texttt{Textures}~\cite{cimpoi_2014_textures}, \texttt{SVHN}~\cite{netzer_2011_SVHN}, \texttt{LSUN-Crop}~\cite{yu_2015_lsun}, \texttt{LSUN-Resize}~\cite{yu_2015_lsun}, \texttt{Place365}~\cite{zhou_2017_places}, and \texttt{iSUN}~\cite{xu_2015_iSUN}. The detailed information of datasets is presented in the appendix. We would like to clarify that the CIFAR benchmark are different from those in the earlier work~\cite{hendrycks_2017_baseline,lee_2018_maha,liang_2018_ODIN}. These new OOD datasets are harder than the earlier OOD dataset like Uniform noise. As shown by recent work~\cite{liu_2020_energy,salehi_2021_OODsurvey}, most existing baselines are far not up to perfect OOD detection performance on this new CIFAR benchmark.

\setParDis\noindent\textbf{Evaluation Metrics.} We use the common metrics to measure the quality of OOD detection: (1) the area under the receiver operating characteristic curve (AUROC); (2) the area under the precision-recall curve (AUPR); (3) the false positive rate of OOD samples when the true positive rate of in-distribution samples is at 95\% (FPR95).

\setParDis\noindent\textbf{Training Details.}  ResNet-18~\cite{he_deep_2016} and WideResNet-28-10~\cite{zagoruyko_wrn_2017} are the main backbones for all methods, and the model is trained by momentum optimizer with the initial learning rate of 0.1. We set the momentum to be 0.9 and the weight decay coefficient to be $2\times 10^{-4}$. Models are trained for 200 epochs and the learning rate decays with a factor of 0.1 at 100 and 150 epochs. Other hyper-parameters for the OOD detection methods come from reference code~\cite{Yang2022Openood} and the original work. We apply LTH with learning weight rewind~\cite{frankle_2020_wrewind}. We prune 20\% of the weights in each pruning stage and repeat 9 times (reaching around a sparsity of 90\%). We run each trial 5 times.

\begin{figure}[h]
	\vskip -0.05in
	\centering
	\includegraphics[width=0.75\linewidth]{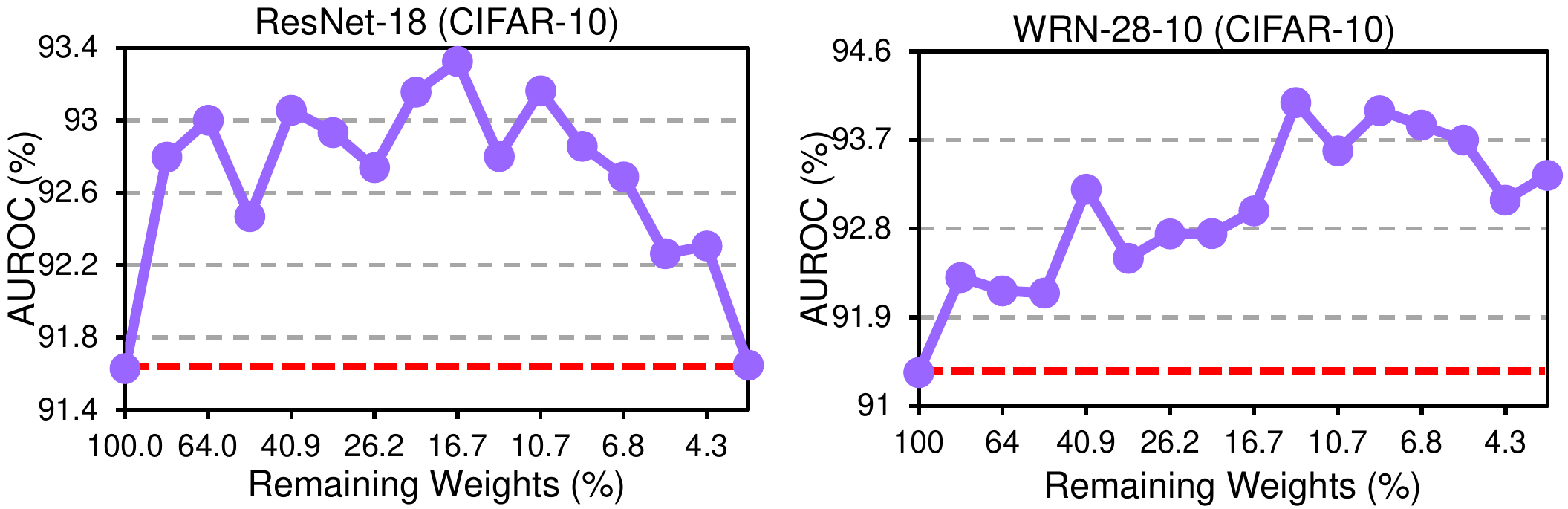}
	\vskip -0.10in
	\caption{Performance on OOD detection (AUROC) based on MSP~\cite{hendrycks_2017_baseline} under different sparsity in AoP. The backbone and training dataset are listed in the title of figure. \textit{We draw the red horizontal line with a value of performance in the dense model}.}
	\label{fig:AoP_base}
\end{figure}

\setParDis\noindent\textbf{AoP boosts OOD detection.} A detailed experiments for comparison with \textit{\textbf{mostly recent methods}} are conducted, and the results are listed in Tab.~\ref{tab:main_results}. AoP could apply to the existing method KNN~\cite{sun2022KNN} and ViM~\cite{wang2022vim}, and boost their performance significantly. \textit{The performance of pruned model based on MSP~\cite{hendrycks_2017_baseline} when we iteratively prune the network} is shown in Figure~\ref{fig:AoP_base}. The models with remaining weights of 100\% mean the dense model without pruning. The performance of pruned models outperforms dense models in a wide range of sparsity. As the sparsity of model increases, the performance of model on OOD detection first rises and then falls. This variation tendency is consistent with intuition since overly pruning the model will make the model lose some important weights.

\begin{figure}[h]
	\vskip -0.15in
	\centering
	\includegraphics[width=0.70\linewidth]{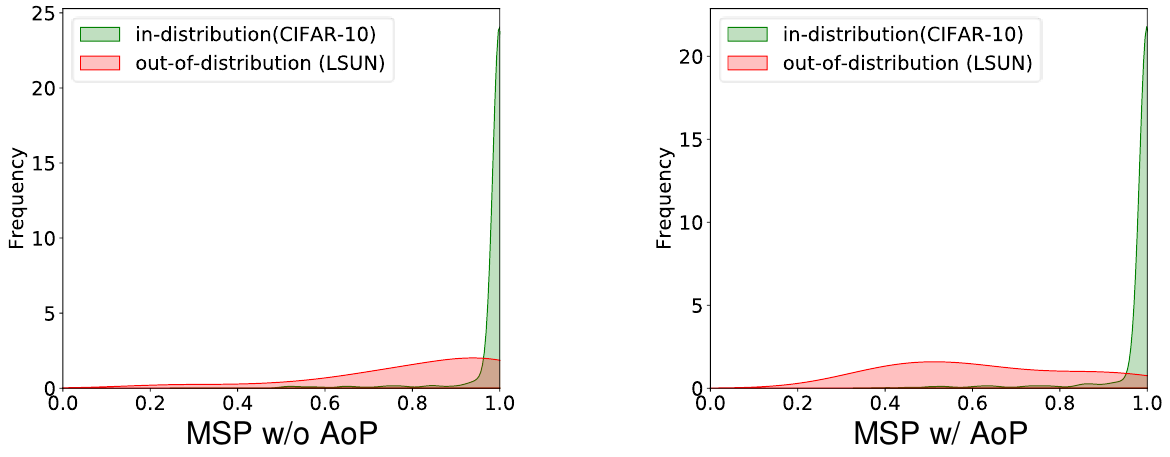}
	\vskip -0.1in
	\caption{Distribution of MSP~\cite{hendrycks_2017_baseline}scores v.s. MSP+AoP scores. The model distinguishes OOD data better while using AoP.}
	\label{fig:AoP_distribution}
	\vskip -0.05in
\end{figure}

\setParDis\noindent\textbf{AoP boosts various OOD scoring functions.} Since AoP is a simple module readily pluggable into the existing methods, we add it to other methods. We report the results of adding AoP to MSP~\cite{hendrycks_2017_baseline} and MaxLogit~\cite{hendrycks2022MaxL}, and it shows consistent improvement in Fig.~\ref{fig:AoP_distribution} and Tab.~\ref{tab:AoP_knn_vim}.

\begin{table}[t]
	\centering
	\caption{Performance on OOD detection under other scoring functions. The model is ResNet-18~\cite{he_deep_2016} trained on CIFAR-10~\cite{krizhevsky2009learning}.}
	\label{tab:AoP_knn_vim}
	\vskip -0.10in
	\renewcommand{\arraystretch}{1.1}
	\renewcommand\tabcolsep{4.0pt}
	\scalebox{0.90}{
		\begin{tabular}{ccccccc}
			\toprule[1.4pt]
			\multirow{2}{*}{Method} & \multicolumn{3}{c}{MSP}    & \multicolumn{3}{c}{MaxLogit}                                     \\ \cmidrule(r){2-4}  \cmidrule(r){5-7} 
			& \multicolumn{1}{c}{AUROC} & \multicolumn{1}{c}{AUPR} & FPR95 & \multicolumn{1}{c}{AUROC} & \multicolumn{1}{c}{AUPR} & FPR95 \\  \cmidrule(r){1-4}  \cmidrule(r){5-7}  
			w/ o AoP   & 91.63    & 98.08     &  50.00   & 92.61      & 98.13   & 35.09    \\ 
			w / AoP    & \textbf{93.32} \footnotesize {\textcolor{magenta}{$\uparrow$ \textbf{1.7}}}  & \textbf{98.64}    &  \textbf{45.95} \footnotesize {\textcolor{magenta}{$\downarrow$ \textbf{4.0}}}  &  \textbf{95.53} \footnotesize {\textcolor{magenta}{$\uparrow$ \textbf{2.9}}}   & \textbf{99.06}   &  \textbf{25.80} \footnotesize {\textcolor{magenta}{$\downarrow$ \textbf{9.3}}}  \\ \toprule[1.4pt]
	\end{tabular}}
\end{table}

\begin{table}[h]
	\centering
	\vskip -0.10in
	\caption{Performance on OOD detection with OE~\cite{hendrycks_2019_oe}. The model is ResNet-18~\cite{he_deep_2016} trained on CIFAR-10~\cite{krizhevsky2009learning}.}
	\label{tab:AoP_oe}
	\vskip -0.10in
	\renewcommand{\arraystretch}{1.1}
	\renewcommand\tabcolsep{13.5pt}
	\scalebox{0.95}{
		\begin{tabular}{ccccc}
			\toprule[1.4pt]
			Method & AUROC & AUPR & FPR95 & Acc \\ \hline
			OE     &  98.66     & 99.73     & 4.49    & 93.92    \\
			\rowcolor{gray!30}
			$\quad$ + AoP & \textbf{98.91}  &  \textbf{99.78}  &  \textbf{3.85}   &   \textbf{94.00} 
			\\ \toprule[1.4pt]
	\end{tabular}}
	\vskip -0.20in
\end{table}

\noindent\textbf{AoP is compatible with OE.} OE~\cite{hendrycks_2019_oe} leverages a dataset of known outliers during training, which is \textit{a very strong baseline in OOD detection}. A recent study~\cite{bitterwolf2022breaking} finds that many methods using auxiliary datasets are equivalent to OE, so we just use OE as the baseline on the behalf of these methods that use extra data. The outlier exposure dataset used in this experiment is 300K Random Images~\cite{hendrycks_2019_oe}. The performance is listed in Tab.~\ref{tab:AoP_oe}. The results show that AoP can achieve consistent improvement on OE.

\begin{figure}[h]
	\vskip -0.10in
	\centering
	\includegraphics[width=0.75\linewidth]{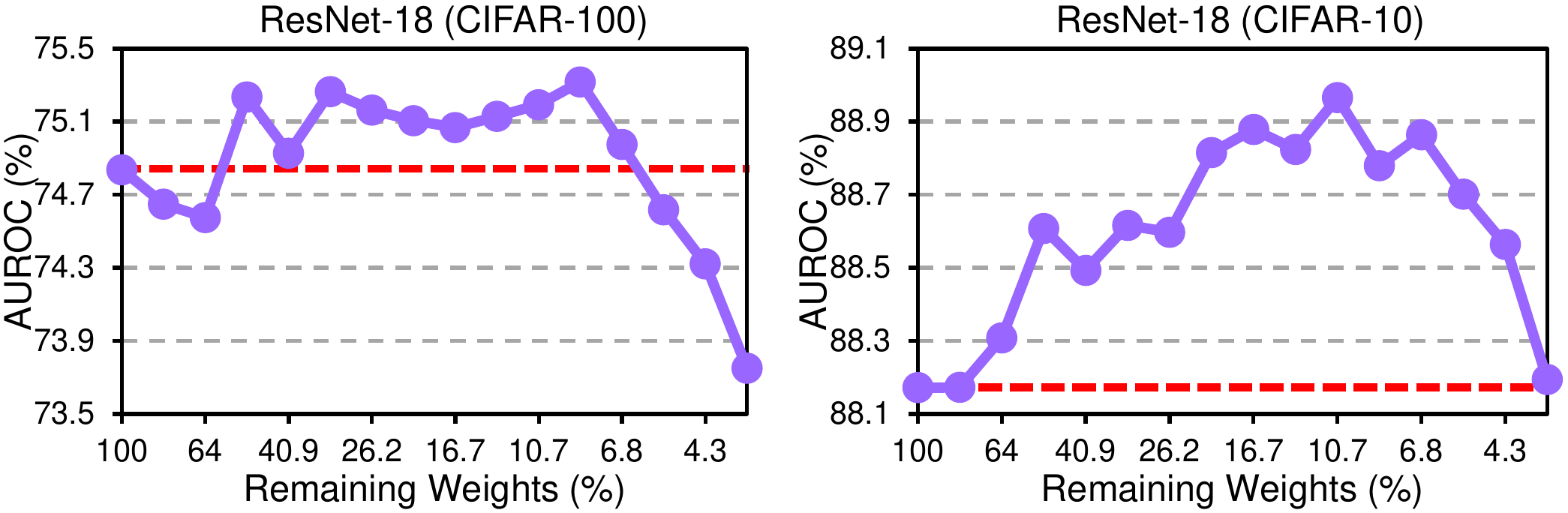}
	\vskip -0.1in
	\caption{Performance of near OOD detection. In-distribution data is CIFAR-100 and OOD data is CIFAR-10, and vice versa. Backbone used is ResNet-18~\cite{he_deep_2016}.}
	\label{fig:AoP_nearOOD}
	\vskip -0.15in
\end{figure}

\noindent\textbf{AoP is beneficial for near-OOD detection tasks.} The difficulty of detecting OOD inputs relies on how semantically close the unknowns are to the inlier classes, which contributes to \textit{near-OOD} task and \textit{far-OOD} task. For example, for a model trained on CIFAR-10, it is hard to detect inputs from CIFAR-100 than SVHN, and the former is regarded as a near-OOD task. A recent study~\cite{salehi_2021_OODsurvey} reveals that none of the methods are good at detecting both near and far OOD samples except OE~\cite{hendrycks_2019_oe}. As shown in Fig.~\ref{fig:AoP_nearOOD}, AoP is helpful in near-OOD task.

\subsection{Evaluation on ImageNet Benchmarks}

\noindent\textbf{In-distribution Datasets.} We use Tiny-ImageNet and ImageNet~\cite{deng2009imagenet} as the in-distribution dataset. Tiny-ImageNet is a 200-class subset of ImageNet where images are resized and cropped to 64$\times$64 resolution. The training set has 100,000 images and the test set has 10,000 images.

\noindent\textbf{Out-of-distribution Datasets.} For the OOD datasets, we use the common benchmarks used in~\cite{liu_2020_energy}: \texttt{Textures}~\cite{cimpoi_2014_textures}, \texttt{iSUN}~\cite{xu_2015_iSUN}, \texttt{LSUN-Resize}~\cite{yu_2015_lsun}, and \texttt{Place365}~\cite{zhou_2017_places}.

\noindent\textbf{Training Details.} For Tiny-ImageNet, the settings are the same as CIFAR benchmarks. For ImageNet, we use one-shot pruning to achieve the sparsity of 80\%, and other settings are the same as~\cite{liu2021GraNet}.

\noindent\textbf{Results.} The results for Tiny-ImageNet are listed in Tab.~\ref{tab:tiny_resnet18}. It reveals that the model with AoP achieves great improvement in OOD detection on three datasets. We also provide the results on ImageNet in Tab.~\ref{tab:AoP_imagenet}. These three OOD datasets follow MOS~\cite{huang_2021_mos}. The results show AoP achieves consistent improvement.

\begin{table}[t]
	\centering
	\caption{OOD detection performance on Tiny-ImageNet using ResNet-18~\cite{he_deep_2016}. The best results are boldfaced for highlight.}
	\label{tab:tiny_resnet18}
	\vskip -0.10in
	\renewcommand{\arraystretch}{1.00}
	\renewcommand\tabcolsep{8.5pt}
	\scalebox{0.92}{
		\begin{tabular}{cccccccccc}
			\toprule[1.2pt]
			\multirow{2}{*}{\textbf{Method}} & \multicolumn{3}{c}{Textures}                                  & \multicolumn{3}{c}{LSUN}  &
			\multicolumn{3}{c}{iSUN}                              \\ \cmidrule(r){2-4} \cmidrule(r){5-7}  \cmidrule(r){8-10}
			& {\begin{tabular}[c]{@{}c@{}}AUROC \end{tabular}} 
			& {\begin{tabular}[c]{@{}c@{}}AUPR \end{tabular}}
			& {\begin{tabular}[c]{@{}c@{}}FPR95 \end{tabular}}  
			& {\begin{tabular}[c]{@{}c@{}}AUROC\end{tabular}}
			& {\begin{tabular}[c]{@{}c@{}}AUPR \end{tabular}}
			& {\begin{tabular}[c]{@{}c@{}}FPR95 \end{tabular}}
			& {\begin{tabular}[c]{@{}c@{}}AUROC \end{tabular}} 
			& {\begin{tabular}[c]{@{}c@{}}AUPR \end{tabular}}
			& {\begin{tabular}[c]{@{}c@{}}FPR95 \end{tabular}}
			\\
			\hline
			MSP~\cite{hendrycks_2017_baseline} &      
			68.32   & 91.87   & 89.17   & 63.79  & 90.89 & 93.8 & 64.76 & 91.14 &92.71    \\
			\rowcolor{gray!30}
			$\quad$+ AoP            &
			\textbf{71.61}   &  \textbf{93.08}  & \textbf{86.35}   & \textbf{66.85}   & \textbf{91.82}  & \textbf{91.77} & \textbf{66.29} & \textbf{91.47} & \textbf{91.20}      \\
			\cmidrule(r){1-4}  \cmidrule(r){5-7} \cmidrule(r){8-10} 
			ODIN~\cite{hendrycks_2017_baseline} &      
			69.03   & 90.08   & 87.45   & 57.87   & 88.70  & 94.22 & 61.22 & 89.49 & 91.46     \\
			\rowcolor{gray!30}
			$\quad$+ AoP            &
			\textbf{74.02}   &  \textbf{93.12}  & \textbf{81.98}   & \textbf{61.65}   & \textbf{90.07}  & \textbf{93.18} & \textbf{63.55} & \textbf{90.17} & \textbf{90.10}      \\
			\cmidrule(r){1-4}  \cmidrule(r){5-7} \cmidrule(r){8-10} 
			Energy~\cite{hendrycks_2017_baseline} &      
			69.01   & 90.81   & 86.09   & 57.25   & 88.51  &93.75 & 60.85 & 89.36 & 91.25     \\
			\rowcolor{gray!30}
			$\quad$+ AoP            &
			\textbf{74.13}   &  \textbf{93.12}  & \textbf{80.16}   & \textbf{60.94}   & \textbf{89.88}  & \textbf{92.45} & \textbf{63.23} & \textbf{90.07} & \textbf{89.38}      \\
			\toprule[1.2pt]
	\end{tabular}}
	\vskip -0.05in
\end{table}

\begin{table}[h]
	\centering
	\footnotesize
	\caption{Performance of OOD detection on ImageNet~\cite{deng2009imagenet} using ResNet-50~\cite{he_deep_2016}.}
	\label{tab:AoP_imagenet}
	\vskip -0.10in
	\renewcommand{\arraystretch}{1.05}
	\renewcommand\tabcolsep{9.0pt}
	\scalebox{1.00}{
		\begin{tabular}{c|cc|cc|cc|cc|cc}
			\toprule[1.4pt]
			Dataset   & MSP & \cellcolor{gray!30}{+ AoP} & ODIN & \cellcolor{gray!30}{+ AoP} & Energy & \cellcolor{gray!30}{+ AoP} & MaxLogit & \cellcolor{gray!30}{+ AoP} & ViM & \cellcolor{gray!30}{+ AoP} \\
			\cmidrule(r){1-3} \cmidrule(r){4-5} \cmidrule(r){6-7} \cmidrule(r){8-9} \cmidrule(r){10-11}
			Places   & 76.26  & 76.58  &  72.91    & 73.90   & 68.41  & 68.49   & 73.73  & 76.40  & 84.74 & 84.91     \\ 
			SUN      & 77.58  & 78.24  & 74.10     & 75.52   & 69.68  & 70.32   & 74.75  & 76.91  & 86.87 & 87.59     \\ 
			Textures & 78.58  & 78.91  &  75.88    & 76.86   & 73.90  & 74.97   & 77.80  & 79.32  & 98.27 & 98.73    \\
			\toprule[1.4pt]
	\end{tabular}}
	\vskip -0.10in
\end{table}

\subsection{Ablation Studies}

\noindent\textbf{Effectiveness of MA and pruning.} To verify the importance of the two components in AoP, we test the performance of OOD detection when they are not used. The results are listed in Tab.~\ref{tab:effectiveness_ablation}. As shown in the table, both of them are essential in OOD detection.

\begin{table}[h]
	\centering
	\caption{Effectiveness of MA and pruning. The in-distribution dataset is CIFAR-10.}
	\label{tab:effectiveness_ablation}
	\vskip -0.10in
	\renewcommand{\arraystretch}{1.1}
	\renewcommand\tabcolsep{5.0pt}
	\scalebox{0.80}{
		\begin{tabular}{cccccccc}
			\toprule[1.4pt]	
			\multirow{2}{*}{MA} & \multirow{2}{*}{pruning} & \multicolumn{3}{c}{ResNet-18}                                     & \multicolumn{3}{c}{WideResNet-28-10}                                     \\ \cmidrule(r){3-5}  \cmidrule(r){6-8}
			&   & \multicolumn{1}{c}{AUROC} & \multicolumn{1}{c}{AUPR} & FPR & \multicolumn{1}{c}{AUROC} & \multicolumn{1}{c}{AUPR} & FPR \\ \cmidrule(r){1-5} \cmidrule(r){6-8} 
			\XSolidBrush &  \XSolidBrush  & \multicolumn{1}{c}{91.63}      & \multicolumn{1}{c}{98.08}     & 50.00    & \multicolumn{1}{c}{91.24}   & \multicolumn{1}{c}{97.72}     &  46.21   \\ 
			\XSolidBrush &   \ding{52}      & \multicolumn{1}{c}{92.94}      & \multicolumn{1}{c}{98.51}     &  46.19   & \multicolumn{1}{c}{93.43}      &  
			\multicolumn{1}{c}{98.22}     & 41.55    \\ 
			\ding{52}&   \ding{52}    & \textbf{93.32}      & \textbf{98.64}   &  \textbf{45.95}   & \textbf{93.99}    & \textbf{98.66}   & \textbf{37.22}    \\ 	\toprule[1.4pt]	 
	\end{tabular}}
	\vskip -0.10in
\end{table}

\begin{table}[h]
	\centering
	\caption{Effectiveness of the three during training pruning techniques. The model is a ResNet-18~\cite{he_deep_2016} trained on CIFAR-10.}
	\label{tab:during_training_pruning}
	\vskip -0.10in
	\renewcommand{\arraystretch}{1.05}
	\renewcommand\tabcolsep{8.0pt}
	\scalebox{0.95}{
		\begin{tabular}{ccccccc}
			\toprule[1.2pt]
			\multirow{2}{*}{Methods} & \multicolumn{2}{c}{Sparsity: 60\%} & \multicolumn{2}{c}{Sparsity: 80\%} & \multicolumn{2}{c}{Sparsity: 90\%} \\  \cmidrule(r){2-3} \cmidrule(r){4-5} \cmidrule(r){6-7}
			& \multicolumn{1}{c}{AUROC}  & Acc  & \multicolumn{1}{c}{AUROC}  & Acc  & \multicolumn{1}{c}{AUROC}  & Acc  \\ \hline
			\multicolumn{1}{r}{GMP~\cite{zhu2017GMP}}   & \multicolumn{1}{c}{92.13} &   94.88   & \multicolumn{1}{c}{92.54} &  94.33    & \multicolumn{1}{c}{92.53}       & 94.32     \\ 
			\multicolumn{1}{r}{RigL~\cite{evci2020rigl}} & \multicolumn{1}{c}{92.66}       &  94.25    & \multicolumn{1}{c}{92.37}       &  93.95    & \multicolumn{1}{c}{92.51}       & 94.08    \\
			\multicolumn{1}{r}{GraNet~\cite{liu2021GraNet}}  & \multicolumn{1}{c}{93.14}       &   94.59   & \multicolumn{1}{c}{92.75}       & 94.54     & \multicolumn{1}{c}{92.99}       &  94.41    \\
			\toprule[1.2pt]
	\end{tabular}}
	\vskip -0.10in
\end{table}

\begin{figure}[h]
	\centering
	\includegraphics[width=0.65\linewidth]{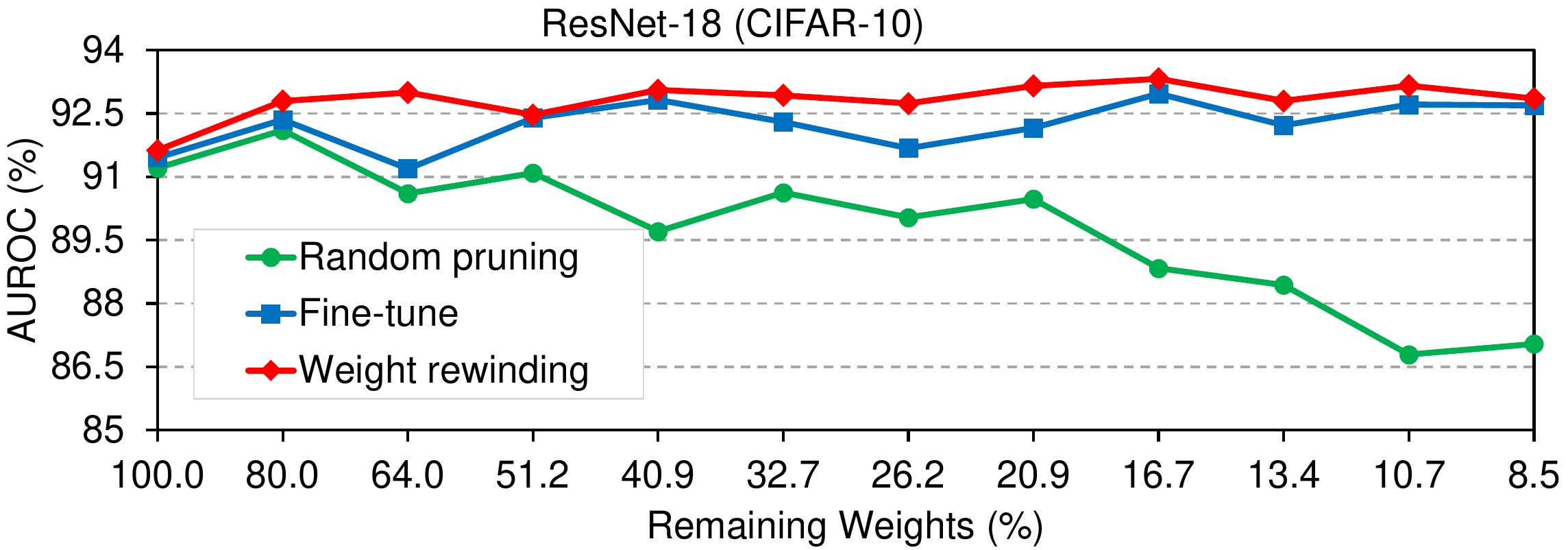}
	\vskip -0.05in
	\caption{Influence of different iterative pruning strategies. The model is a ResNet-18~\cite{he_deep_2016} trained on CIFAR-10.}
	\label{fig:AoP_pruning_techique}
	\vskip -0.10in
\end{figure}

\noindent\textbf{Influence of pruning techniques.} \textit{(1) Different iterative pruning methods.} LTH is a kind of iterative pruning~\cite{frankle_2018_LTH,frankle_2020_wrewind}. We compare it with other two iterative pruning strategies: random pruning and fine-tuning~\cite{frankle_2018_LTH}. The results on CIFAR-10 are shown in Fig.~\ref{fig:AoP_pruning_techique}. It shows that LTH with weight rewind achieves the best performance. The performance of fine-tuning is poorer, verifying that the initialization of weights is important in OOD detection. Random pruning decreases the performance, revealing that the weights can not be dropped arbitrarily. \textit{(2) During-training pruning methods.}

\begin{wrapfigure}{l}{0.30\textwidth}
	\centering
	\vskip -0.15in
	\includegraphics[width=0.30\textwidth]{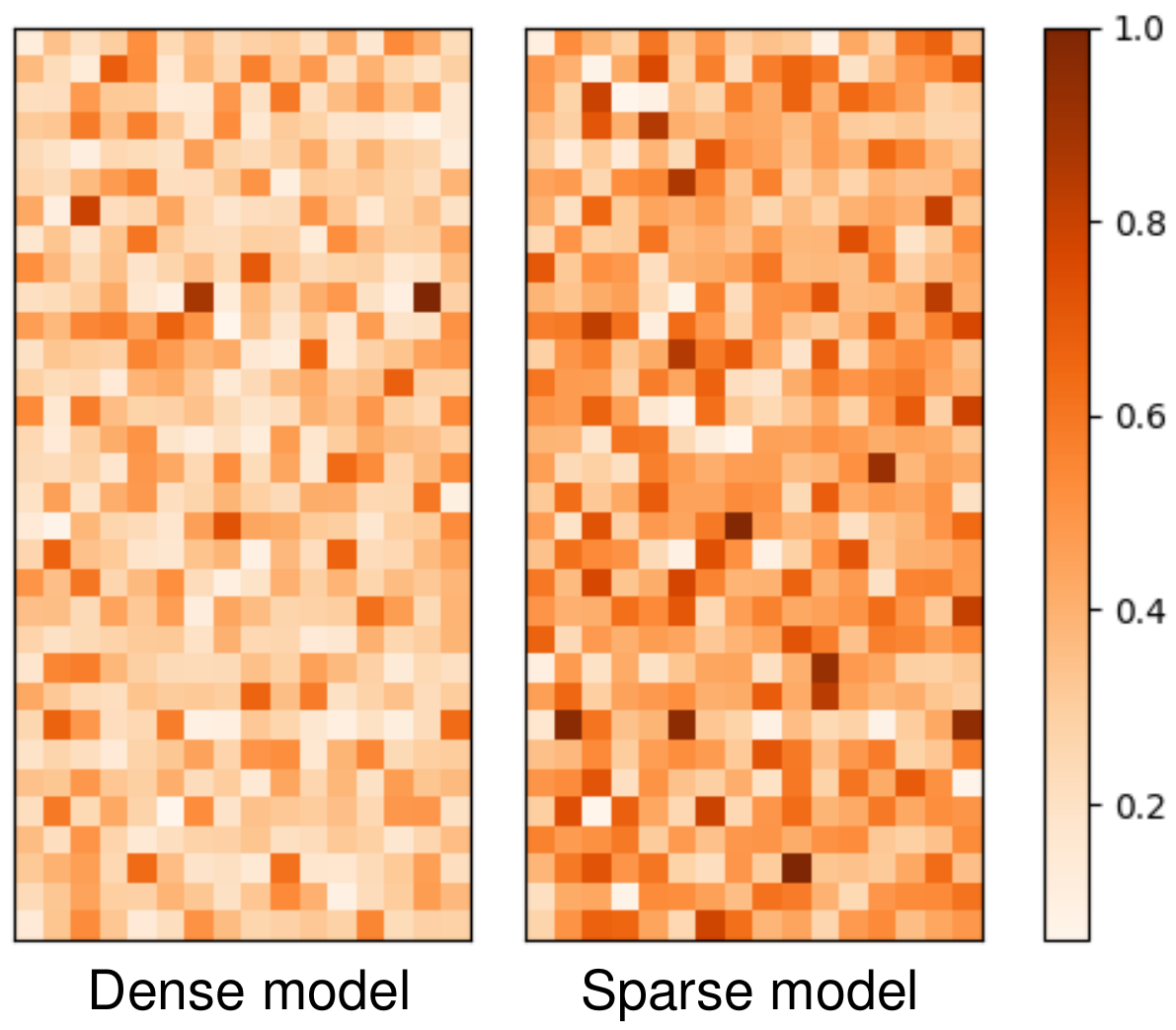}
	\caption{The difference of extracted feature between ID and OOD data. Features extracted by dense model seem to be similar, while it is distinct for sparse model.} 
	\label{fig:AoP_features}
	\vskip -0.25in
\end{wrapfigure}
Since LTH is a kind of post-training pruning, we also explore whether during-training pruning is beneficial for OOD detection. We show the performance of three methods: GMP~\cite{zhu2017GMP}, RigL~\cite{evci2020rigl}, and GraNet~\cite{liu2021GraNet}. The settings about pruning mainly follow GraNet~\cite{liu2021GraNet}. The results are listed in Tab.~\ref{tab:during_training_pruning}. These pruning techniques are good for OOD detection, while they are poorer than LTH~\cite{frankle_2020_wrewind}.


\noindent\textbf{Effectiveness on other networks.} We conduct more experiments on other networks to verify that AoP promotes OOD detection, including Vgg-16~\cite{simonyan_2014_vgg}, WideResNet-40-2~\cite{zagoruyko_wrn_2017} and MobileNet~\cite{howard_2017_mobilenets}. AoP achieves consistent promotion. To save space, results are shown in the appendix.

\setParDis\noindent\textbf{Visualization of features.} To show the influence of AoP on feature representation in OOD detection, we visualize the final features. We randomly sample a batch of data from CIFAR-10~\cite{krizhevsky2009learning} (ID data) and LSUN~\cite{yu_2015_lsun} (OOD data), and calculate the difference of feature representations for dense models. We also calculate the difference for sparse model. We visualize the absolute value of difference of final features. The results are shown in Fig.~\ref{fig:AoP_features}. In the figure, most of the features from ID or OOD dataset are similar for dense model. However, their difference are more obvious for sparse model. It verifies that the sparsity drives model to drop common features existing in both ID and OOD datasets.

\subsection{AoP Improves Misclassification Detection}

\noindent\textbf{Background.} We investigate the effect of AoP on Misclassification Detection (MD). OOD detection and MD both reflect whether models know what they do not know. In safety-critical settings, the classifier should output lower confidence for both incorrect predictions from known classes and OOD examples from unknown (new) classes. The benchmarks of these two tasks are proposed and studied together in~\cite{hendrycks_2017_baseline,moon_2020_confidence}. We study these two tasks to provide a more comprehensive evaluation, which can better verify the effect of our findings.

\setParDis\noindent\textbf{Setups.} The hyper-parameters for models are the same as settings in Sec.~\ref{sec:cifar_benchmarks}. The metrics include AUROC, AURC, E-AURC, and FPR95. We provide a detailed description of these metrics in the supplementary material. We use softmax (i.e., MSP~\cite{hendrycks_2017_baseline}) and CRL~\cite{moon_2020_confidence} as baselines.

\begin{figure}[h]
	\vskip -0.05in
	\centering
	\includegraphics[width=0.75\linewidth]{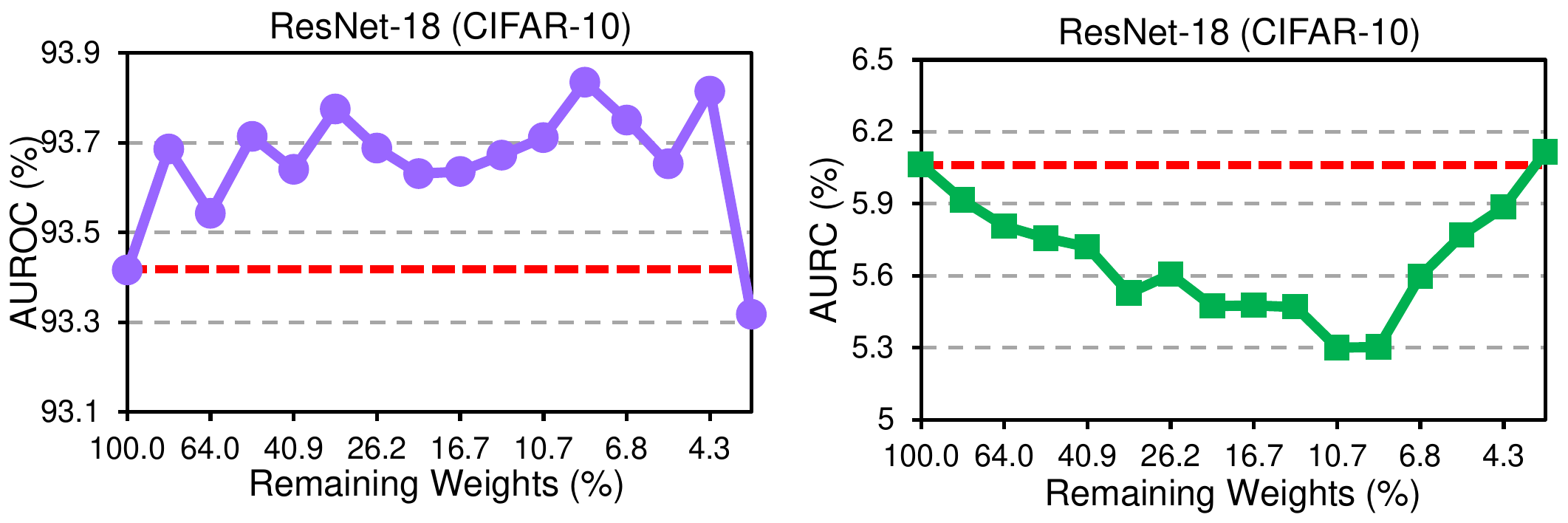}
	\vskip -0.1in
	\caption{Evaluation of misclassified examples detection under two metrics: AUROC and AURC. Backbone is ResNet-18~\cite{he_deep_2016}.}
	\label{fig:AoP_md}
	\vskip -0.25in
\end{figure}

\setParDis\noindent\textbf{Results.} These results in Tab.~\ref{tab:cc_md} reveal that AoP also has a significant improvement in all metrics. We provide the curve of AUROC and FPR95 in Fig.~\ref{fig:AoP_md}. The curves show that AoP can significantly improve the performance of misclassification detection. More results on other networks and datasets are provided in the appendix.

\begin{table}[t]
	\centering
	\caption{Performance of misclassified detection in CIFAR-10 using WideResNet-40-2~\cite{zagoruyko_wrn_2017}. AURC and E-AURC values are multiplied by $10^3$. Other values are percentages. The best results are boldfaced for highlight.}
	\label{tab:cc_md}
	\renewcommand{\arraystretch}{1.01}
	\renewcommand\tabcolsep{5.0pt}
	\scalebox{0.95}{
		\begin{tabular}{cccccc}
			\toprule[1.2pt]
			Method & AUROC $\uparrow$  & AURC $\downarrow$ & E-ARUC $\downarrow$ & AUPR-Err $\uparrow$ & FPR95 $\downarrow$ 
			\\ \hline
			MSP~\cite{hendrycks_2017_baseline}   & 92.51      &  7.55   &  5.65      &  44.56       &   40.87    
			\\ \rowcolor{gray!30}
			$\quad$ + AoP &   \textbf{93.23}    &  \textbf{6.41}   &   \textbf{4.64}    & \textbf{45.79}  &  \textbf{38.47}     
			\\  \hline
			CRL~\cite{moon_2020_confidence} & 93.58      &  6.63   &  4.60      &  45.94   & 40.03     
			\\ \rowcolor{gray!30}
			$\quad$ + AoP &  \textbf{93.98}  & \textbf{6.00} &  \textbf{4.09} & \textbf{47.81} & \textbf{36.07}   
			\\ \toprule[1.2pt]
	\end{tabular}}
	\vskip -0.20in
\end{table}

\section{Conclusion}

In this work, we investigate the chaotic behavior of OOD detection along the optimization trajectory during training, which is ignored by previous studies. We find OOD detection suffers from instability and overfitting, and verify that AoP, consisting of model averaging and pruning, can eliminate such two drawbacks. Model average smooth the weights of models, contributing to a stable performance of the final checkpoint. Pruning increases the sparsity of the model, evading learning too many common features. We also provide a theoretical analysis of the connection between sparsity and OOD detection. The comprehensive experiments verify that AoP achieves great improvement on OOD detection.

\bibliographystyle{plainnat}
\bibliography{reference}

\appendix
\onecolumn
\clearpage
\begin{center}{\bf {\LARGE Supplementary Material:}}
\end{center}
\begin{center}{\bf {\LARGE Average of Pruning: Improving Performance and Stability of Out-of-Distribution Detection}}
\end{center}
\vspace{0.1in}

This appendix can be divided into 6 parts. To be precise,
\begin{itemize}
	\item In Section~\ref{sec:dataset}, we give detailed descriptions of the OOD datasets.
	\item In Section~\ref{sec:metrics}, we give some details about the evaluation metrics used in the main text.
	\item In Section~\ref{sec:proofs}, we present the proofs of the theorems in the main text.
	\item In Section~\ref{sec:configurations}, we show the experimental configurations in detail.
	\item In Section~\ref{sec:addtional_results}, we provide some additional experimental results that we omit in the main body of the paper because of the limitation in space.
	\item In Section~\ref{sec:future_work}, we provide some discussions about our work.
\end{itemize}

\section{Dataset Details}
\label{sec:dataset}

We would like to introduce the OOD datasets used in the paper.
\begin{itemize}
	\item [$\bullet$] \textbf{Textures~\cite{cimpoi_2014_textures}} is an evolving collection of textural images in the wild, consisting of 5640 images, organized into 47 items.
	\item [$\bullet$] \textbf{SVHN~\cite{netzer_2011_SVHN}} contains 10 classes comprised of the digits 0-9 in street view, which contains 26,032 images for testing.
	\item [$\bullet$] \textbf{Place365}~\cite{zhou_2017_places} consists in 1,803,460 large-scale photographs of scenes. Each photograph belongs to one of 365 classes. Following the work of~\cite{hendrycks_2019_oe,liu_2020_energy}, we use a subset of Place365 as OOD data.
	\item [$\bullet$] \textbf{LSUN~\cite{yu_2015_lsun}} has a test set of 10,000 images of 10 different scene categories. Following~\cite{liang_2018_ODIN,liu_2020_energy}, we construct two datasets, \textit{LSUN-crop} and \textit{LSUN-resize}, by randomly cropped and downsampling LSUN test set, respectively.
	\item [$\bullet$] \textbf{iSUN~\cite{xu_2015_iSUN}} is a ground truth of gaze traces on images from the SUN dataset. The dataset contains 2,000 images for the test.
\end{itemize}

\section{Evaluation Metrics}
\label{sec:metrics}
We would like to introduce the metrics applied in the main text.
\subsection{Metrics in out-of-distribution detection}
\begin{itemize}
	\item [$\bullet$] \textbf{AUROC} measures the Area Under the Receiver Operating Characteristic curve (AUROC). The ROC curve depicts the relationship between True Positive Rate and False Positive Rate.
	\item [$\bullet$] \textbf{AUPR} is the Area under the Precision-Recall (PR) curve. The PR curve is a graph showing the precision=TP/(TP+FP) versus recall=TP/(TP+FN). AUPR typically regards the in-distribution samples as positive samples.
	\item [$\bullet$] \textbf{FPR-95\%-TPR} (FPR95) can be interpreted as the probability that a negative (OOD data) sample is classified as an in-distribution prediction when the true positive rate (TPR) is as high as 95\%. We denote it as FPR95 for short.
\end{itemize}

\subsection{Metrics in misclassification detection}
The metrics of AUROC and FPR95 have been introduced in the previous part, so we give descriptions of the remaining metrics: AURC, E-AURC, and AUPR-Err. The difference of AUROC and FPR95 in OOD detection and misclassification detection is that the former regards OOD data as negatives while the latter regards misclassified examples as negatives.
\begin{itemize}
	\item [$\bullet$] \textbf{AURC} is the area under the risk-coverage curve. Risk-coverage curve is defined by the work of~\cite{geifman2018biasreduced}.
	\item [$\bullet$] \textbf{E-AURC}, known as excess AURC, is a normalization of AURC where we subtract the AURC of the best score function in hindsight~\cite{geifman2018biasreduced}.
	\item [$\bullet$] \textbf{AUPR-Err} is the Area under the Precision-Recall (PR) curve. The PR curve is a graph showing the precision versus recall. The metric AUPR-Err indicates the area under the precision-recall curve where errors are specified as positives, since we want to detect misclassified examples.
\end{itemize}

\section{Proofs}
\label{sec:proofs}

\subsection{Proof for Theorem 1}
\begin{theorem}\label{pro:simplified_risk}
	Suppose the distributions of ID and OOD data follow Gaussian distribution in Eq.~(2), then $\mathcal{R} _{\mathrm{ID}}\left( f_{\mathrm{Bayes}} \right)$ and $\mathcal{R} _{\mathrm{OOD}}\left( f_{\mathrm{Bayes}} \right)$ could be simplified to
	\begin{equation}\label{equ:risk_simplified}\nonumber
		\begin{aligned}
			&\mathcal{R} _{\mathrm{ID}}\left( f_{\mathrm{Bayes}} \right) =\,\,\mathrm{Pr}\left\{ \mathcal{N} \left( 0,1 \right) >\frac{\sqrt{1+d\eta ^2}}{\sigma} \right\},
			\\[10pt]
			&\mathcal{R} _{\mathrm{OOD}}\left( f_{\mathrm{Bayes}} \right) =\mathrm{Pr}\left\{ \mathcal{N} \left( 0,1 \right) >\frac{\delta -d\eta ^2}{\sigma \cdot \sqrt{1+d\eta ^2}} \right\} 
			+\mathrm{Pr}\left\{ \mathcal{N} \left( 0,1 \right) >\frac{\delta +d\eta ^2}{\sigma \cdot \sqrt{1+d\eta ^2}} \right\}.
		\end{aligned}
	\end{equation}
\end{theorem}

\noindent \textit{Proof.} Under the assumption of distribution for ID and OOD data in Eq.~(2), the decision is based on the sign of logits from Bayes classifier:
\begin{equation}\label{key}\nonumber
	f_{\mathrm{Bayes}}\left( \boldsymbol{x} \right) =x_1+\eta \sum_{i=2}^{d+1}{x_i}.
\end{equation}

\noindent The in-distribution classification risk of $f_{\mathrm{Bayes}}$, i.e., $\mathcal{R} _{\mathrm{ID}}\left( f_{\mathrm{Bayes}} \right)$, is
\begin{equation}\label{key}\nonumber
	\begin{aligned}
		\mathcal{R} _{\mathrm{ID}}\left( f_{\mathrm{Bayes}} \right) &=\underset{\left( \boldsymbol{x},y \right) \in \mathcal{D} _{\mathrm{ID}}}{\mathrm{Pr}}\left\{ \mathrm{sign}\left( f\left( \boldsymbol{x} \right) \right) \ne y \right\} 
		\\[10pt]
		&=\underset{\left( \boldsymbol{x},y \right) \in \mathcal{D} _{\mathrm{ID}}}{\mathrm{Pr}}\left\{ y\cdot \left( \mathcal{N} \left( y,\sigma ^2 \right) +\sum_{i=2}^{d+1}{\eta \mathcal{N} \left( y\eta ,\sigma ^2 \right)} \right) <0 \right\} 
		\\[10pt]
		&=\underset{\left( \boldsymbol{x},y \right) \in \mathcal{D} _{\mathrm{ID}}}{\mathrm{Pr}}\left\{ \mathcal{N} \left( 1,\sigma ^2 \right) +\sum_{i=2}^{d+1}{\eta \mathcal{N} \left( \eta ,\sigma ^2 \right) <0} \right\} 
		\\[10pt]
		&=\underset{\left( \boldsymbol{x},y \right) \in \mathcal{D} _{\mathrm{ID}}}{\mathrm{Pr}}\left\{ \mathcal{N} \left( 1+d\eta ^2,\left( 1+d\eta ^2 \right) \sigma ^2 \right) <0 \right\} 
		\\[10pt]
		&=\mathrm{Pr}\left\{ \mathcal{N} \left( 0,1 \right) >\frac{\sqrt{1+d\eta ^2}}{\sigma} \right\} .
	\end{aligned}
\end{equation}

\noindent The out-of-distribution rejection risk of $f_{\mathrm{Bayes}}$, i.e., $\mathcal{R} _{\mathrm{OOD}}\left( f_{\mathrm{Bayes}} \right)$, is
\begin{equation}\label{key}\nonumber
	\begin{aligned}
		\mathcal{R} _{\mathrm{OOD}}\left( f_{\mathrm{Bayes}} \right) &=\underset{\left( \boldsymbol{x},q \right) \in \mathcal{D} _{\mathrm{OOD}}}{\mathrm{Pr}}\left\{ \left| f\left( \boldsymbol{x} \right) \right|>\delta \right\} 
		\\[10pt]
		&=\underset{\left( \boldsymbol{x},q \right) \in \mathcal{D} _{\mathrm{OOD}}}{\mathrm{Pr}}\left\{ \left( \mathcal{N} \left( 0,\sigma ^2 \right) +\sum_{i=2}^{d+1}{\eta \mathcal{N} \left( \eta ,\sigma ^2 \right)} \right) >\delta \right\}\\
		&\quad \quad \quad+\underset{\left( \boldsymbol{x},q \right) \in \mathcal{D} _{\mathrm{OOD}}}{\mathrm{Pr}}\left\{ \left( \mathcal{N} \left( 0,\sigma ^2 \right) +\sum_{i=2}^{d+1}{\eta \mathcal{N} \left( \eta ,\sigma ^2 \right)} \right) <-\delta \right\} \,\,
		\\[10pt]
		&=\underset{\left( \boldsymbol{x},q \right) \in \mathcal{D} _{\mathrm{OOD}}}{\mathrm{Pr}}\left\{ \mathcal{N} \left( d\eta ^2,\left( 1+d\eta ^2 \right) \sigma ^2 \right) >\delta \right\} \\
		&\quad \quad \quad +\underset{\left( \boldsymbol{x},q \right) \in \mathcal{D} _{\mathrm{OOD}}}{\mathrm{Pr}}\left\{ \mathcal{N} \left( d\eta ^2,\left( 1+d\eta ^2 \right) \sigma ^2 \right) <-\delta \right\} 
		\\[10pt]
		&=\mathrm{Pr}\left\{ \mathcal{N} \left( 0,1 \right) >\frac{\delta -d\eta ^2}{\sigma \sqrt{1+d\eta ^2}} \right\} +\mathrm{Pr}\left\{ \mathcal{N} \left( 0,1 \right) >\frac{\delta +d\eta ^2}{\sigma \sqrt{1+d\eta ^2}} \right\} .
	\end{aligned}
\end{equation}

\subsection{Proof for Theorem 2}
\begin{theorem}\label{pro:LASSO_solution}
	The classifier $f_{\mathrm{LASSO}}$~\cite{hastie2009esl} obtained by minimizing the loss in Eq.~(6) is
	\begin{equation}\label{equ:lasso_classifier}\nonumber
		f_{\mathrm{LASSO}}\left( \boldsymbol{x} \right) = \left( \left( 1-\lambda \right) _+\cdot x_1+\sum_{i=2}^{d+1}{\left( \eta -\lambda \right) _+\cdot x_i} \right),
	\end{equation}
	where $\left( \cdot \right) _+$ means the positive part of a number.
\end{theorem}

\noindent \textit{Proof.} Taking the derivative of loss with LASSO~\cite{hastie2009esl} in Eq.~(6) with respect to $\boldsymbol{w}$ and setting equal to 0 gives
\begin{equation}\label{key}\nonumber
	\mathbb{E} _{\left( \boldsymbol{x},y \right) \sim \mathcal{D} _{\mathrm{ID}}}\left\{ -\frac{\boldsymbol{x}}{\sigma}\left( y-\frac{\boldsymbol{x}^{\mathrm{T}}\boldsymbol{w}}{\sigma} \right) +\lambda \cdot \mathrm{sign}\left( \boldsymbol{w} \right) \right\} =0.
\end{equation}
\noindent Considering the distribution of ID data, then $\mathbb{E} _{\left( \boldsymbol{x},y \right) \sim \mathcal{D} _{\mathrm{ID}}}\frac{\boldsymbol{xx}^{\mathrm{T}}}{\sigma ^2}=I$, and $\mathbb{E} _{\left( \boldsymbol{x},y \right) \sim \mathcal{D} _{\mathrm{ID}}}\left( \frac{\boldsymbol{x}y}{\sigma} \right) =\left[ 1,\eta ,\dots ,\eta \right] ^ \mathrm{T}$.
Then, we could get the solution of LASSO as follows
\begin{equation}\label{key}\nonumber
	\begin{aligned}
		\boldsymbol{w}_{\mathrm{LASSO}}&=\mathbb{E} _{\left( \boldsymbol{x},y \right) \sim \mathcal{D} _{\mathrm{ID}}}\left\{ \frac{\boldsymbol{x}y}{\sigma}-\lambda \cdot \mathrm{sign}\left( \boldsymbol{w}_{\mathrm{LASSO}} \right) \right\} 
		\\[5pt]
		&=\left[ 1,\eta ,\cdots ,\eta \right] ^\mathrm{T}-\lambda \cdot \mathrm{sign}\left( \boldsymbol{w}_{\mathrm{LASSO}} \right) 
		\\[5pt]
		&=\left[ \left( 1-\lambda \right) _+,\left( \eta -\lambda \right) _+,\cdots ,\left( \eta -\lambda \right) _+ \right] ^{\mathrm{T}}.
	\end{aligned}
\end{equation}

\noindent Thus, the output of LASSO is as follows
\begin{equation}\label{key}\nonumber
	f_{\mathrm{LASSO}}\left( \boldsymbol{x} \right) =\boldsymbol{w}_{\mathrm{LASSO}}^{\mathrm{T}}\boldsymbol{x}=\left( 1-\lambda \right) _+\cdot x_1+\sum_{i=2}^{d+1}{\left( \eta -\lambda \right) _+\cdot x_i}.
\end{equation}
\noindent More detailed results about the solution of LASSO could be found in Section 3.4 from the textbook~\cite{hastie2009esl}.

\section{Experimental Configurations}
\label{sec:configurations}
\subsection{Experimental settings on CIFAR benchmarks}
We would like to add some details about the settings of AoP on CIFAR benchmarks. For model averaging, we set the started epoch of the model averaging $t_0$ to 100. For pruning (i.e., LTH with weight rewinding~\cite{frankle_2020_wrewind}), we prune 20\% of the weights in each pruning
stage and repeat 9 times (reaching a sparsity of 86.3\%). \textit{To show the influence of sparsity on OOD detection, we continue to prune the model 15 times, and draw the curve between sparsity and performance of OOD detection.}

\subsection{Experimental settings of other OOD detection methods}
The OOD detection methods compared in the main text include MSP~\cite{hendrycks_2017_baseline}, ODIN~\cite{liang_2018_ODIN}, Mahalanobis distance~\cite{lee_2018_maha}, energy score~\cite{liu_2020_energy}, GradNorm~\cite{huang2021gradnorm}, ReAct~\cite{sun_2021_react}, MaxLogit~\cite{hendrycks2022MaxL}, VOS~\cite{du2022vos}, KNN~\cite{sun2022KNN}, and ViM~\cite{wang2022vim}. Most of our settings refer to the work of \textit{OpenOOD}~\cite{Yang2022Openood} and their original codes.

For ODIN~\cite{liang_2018_ODIN}, we set the temperature scaling $T$ to 1000. For Mahalanobis distance~\cite{lee_2018_maha}, we set the magnitude of noise to 0.005. For energy score, we set the temperature scaling $T$ to 1. For ReAct~\cite{sun_2021_react}, we set the rectification percentile to 90. For KNN~\cite{sun2022KNN}, we set $k$ to 50. For ViM~\cite{wang2022vim}, we set the dimension of principle space to 256.

\section{Additional Experimental Results}
\label{sec:addtional_results}

We would like to provide more experimental results which are not included in the main text because of the limit in space. We provide more curves about the phenomenon of instability and overfitting that we find in the paper.

\subsection{Experiments about instability and overfitting}
\begin{figure}[h]
	\vskip -0.05in
	\centering
	\includegraphics[width=0.98\linewidth]{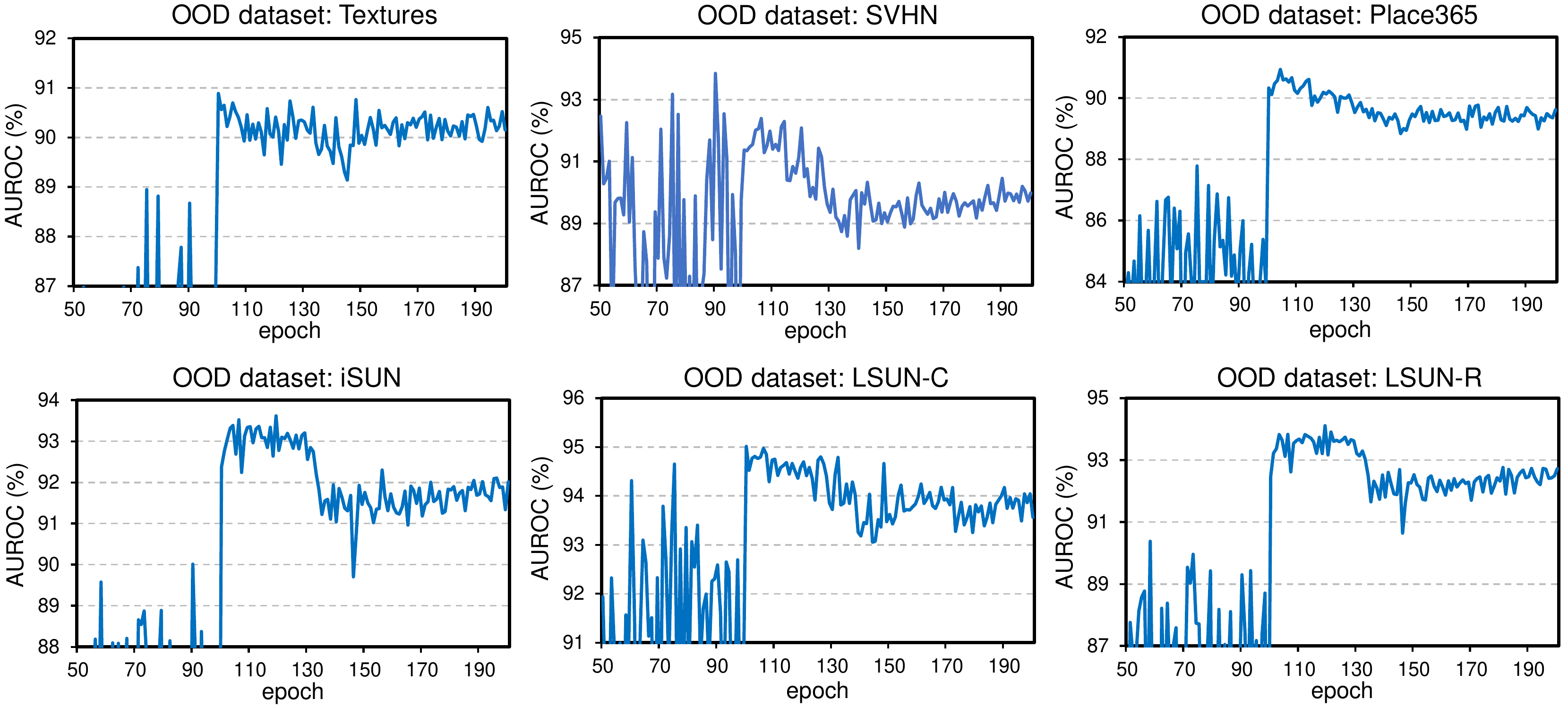}
	\vskip -0.10in
	\caption{Instability and overfitting occur in various OOD datasets. The model is ResNet-18~\cite{he_deep_2016} trained on CIFAR-10. The OOD datasets include Textures~\cite{cimpoi_2014_textures}, SVHN~\cite{netzer_2011_SVHN}, Place365~\cite{zhou_2017_places}, iSUN~\cite{xu_2015_iSUN}, LSUN-C~\cite{yu_2015_lsun}, and LSUN-R~\cite{yu_2015_lsun}. The OOD datasets are indicated in the title of each figure. The results show that instability and overfitting exist in different OOD datasets.}
	\label{fig:app_ood_datasets}
	\vskip -0.10in
\end{figure}

\begin{figure}[h]
	\vskip -0.10in
	\centering
	\includegraphics[width=0.98\linewidth]{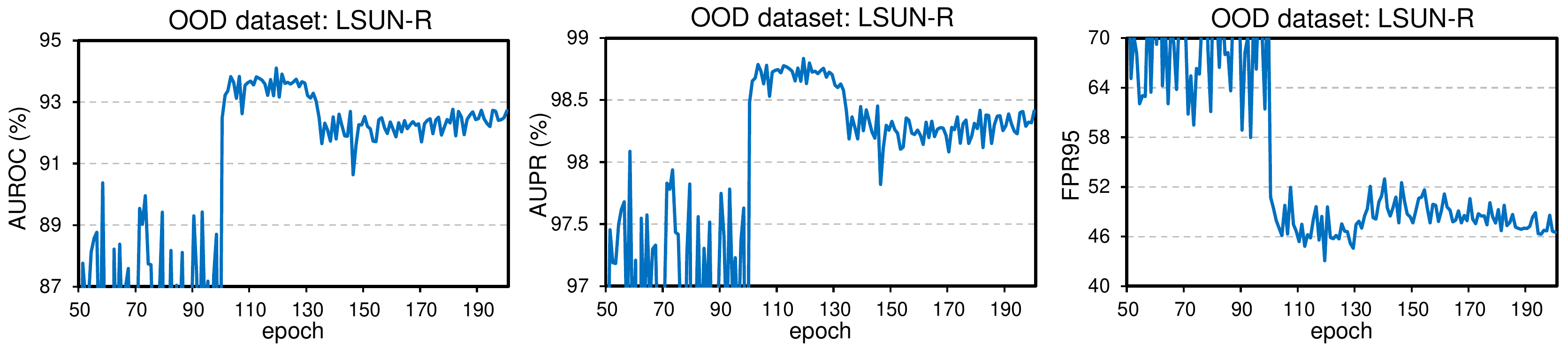}
	\vskip -0.10in
	\caption{Different metrics (AUROC, AUPR, and FPR95) in OOD detection show instability and overfitting. The model is ResNet-18~\cite{he_deep_2016} trained on CIFAR-10. The OOD dataset used is LSUN-R~\cite{yu_2015_lsun}.}
	\label{fig:app_ood_metrics}
	\vskip -0.15in
\end{figure}

\textit{\textbf{The phenomenon of instability and overfitting exists in different OOD datasets.}} In the main text, we show that the curve of the model trained on CIFAR-10 and tested on LSUN-R~\cite{yu_2015_lsun}. The curves of testing on other OOD datasets are shown in Fig.~\ref{fig:app_ood_datasets}. Due to the limit of space, we show the curve of AUROC in the main text. We also show how AUPR and FPR95 vary during the training stage, and the results are shown in Fig.~\ref{fig:app_ood_metrics}. It reveals instability and overfitting exist in various metrics.

\begin{figure}[t]
	\centering
	\includegraphics[width=0.98\linewidth]{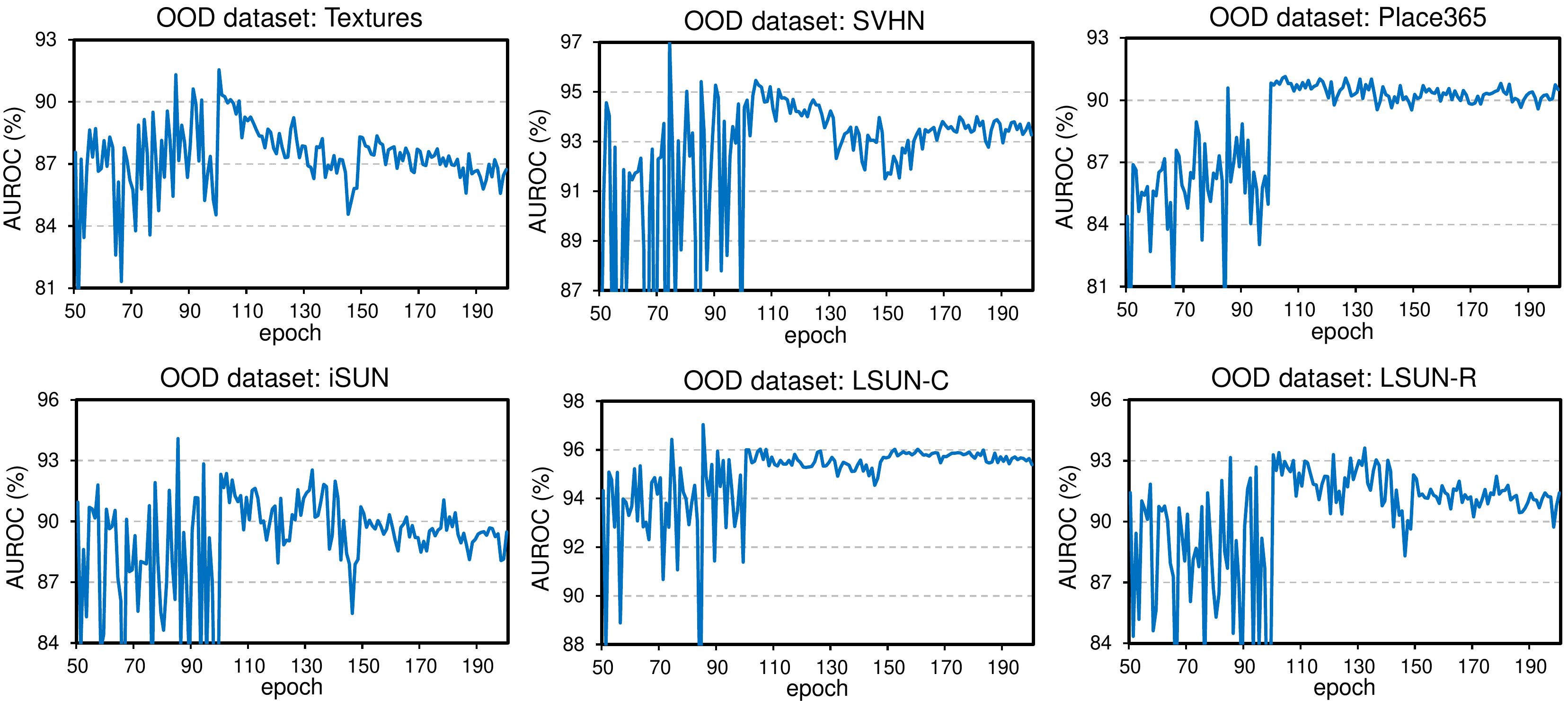}
	\vskip -0.10in
	\caption{Instability and overfitting occur in other network. The model is WideResNet-28-10~\cite{zagoruyko_wrn_2017} trained on CIFAR-10. The OOD datasets include Textures~\cite{cimpoi_2014_textures}, SVHN~\cite{netzer_2011_SVHN}, Place365~\cite{zhou_2017_places}, iSUN~\cite{xu_2015_iSUN}, LSUN-C~\cite{yu_2015_lsun}, and LSUN-R~\cite{yu_2015_lsun}. The OOD datasets are indicated in the title of each figure.}
	\label{fig:app_ind_wrn}
	\vskip -0.20in
\end{figure}

\textit{\textbf{The phenomenon of instability and overfitting occurs in different networks.}} We show that instability and overfitting also exist in other networks, like WideResNet-28-10~\cite{zagoruyko_wrn_2017}, and the curves are presented in Fig.~\ref{fig:app_ind_wrn}.


\subsection{Results of AoP on other networks}

\noindent \textbf{OOD detection}. We conduct more experiments on other networks to verify that AoP promotes OOD detection, including Vgg-16~\cite{simonyan_2014_vgg}, WideResNet-40-2~\cite{zagoruyko_wrn_2017} and MobileNet~\cite{howard_2017_mobilenets}. The results are listed in Tab.~\ref{tab:extra_results}. As shown in the results, AoP achieves consistent improvement in OOD detection.

\begin{table}[h]
	\centering
	\vskip -0.10in
	\caption{Performance of OOD detection for other networks. The models are trained on CIFAR-10. The testing OOD datasets are the same as CIFAR benchmarks in the main text. The best results are boldfaced for highlighting.}
	\vskip -0.10in
	\label{tab:extra_results}
	\renewcommand{\arraystretch}{1.3}
	\renewcommand\tabcolsep{10.0pt}
	\scalebox{0.80}{
		\begin{tabular}{cccccccccc}
			\toprule[1.4pt]	
			\multirow{2}{*}{Method} & \multicolumn{3}{c}{Vgg-16}  & \multicolumn{3}{c}{WRN-40-2}       & \multicolumn{3}{c}{MobileNet}    \\ \cmidrule(r){2-4}   \cmidrule(r){5-7}  \cmidrule(r){8-10}
			& \multicolumn{1}{c}{AUROC $\uparrow$} & \multicolumn{1}{c}{AUPR $\uparrow$} & \multicolumn{1}{c}{FPR $\downarrow$} & \multicolumn{1}{c}{AUROC $\uparrow$} & \multicolumn{1}{c}{AUPR $\downarrow$} & \multicolumn{1}{c}{FPR $\downarrow$} & \multicolumn{1}{c}{AUROC $\uparrow$} &  \multicolumn{1}{c}{AUPR $\uparrow$} & \multicolumn{1}{c}{FPR $\downarrow$}  
			\\  \cmidrule(r){1-4}   \cmidrule(r){5-7}  \cmidrule(r){8-10}
			MSP  & 88.88 & 97.51 & 59.22 &  89.20  & 97.87 & 59.88    & 88.05  & 97.61 &  61.96  \\ 
			\rowcolor{gray!30}
			$\quad$ + AoP  & \textbf{89.88}  & \textbf{97.70} & \textbf{56.63}    & \textbf{90.24}  & \textbf{98.14} & \textbf{57.25}   &  \textbf{89.93}   &  \textbf{98.21}  &  \textbf{58.96}   \\ 
			\toprule[1.4pt]
	\end{tabular}}
\end{table}

\noindent \textbf{Misclassification detection}. We conduct more experiments on other networks to verify that AoP promotes misclassification detection, including Vgg-16~\cite{simonyan_2014_vgg}, and ResNet-18~\cite{he_deep_2016}. The results are listed in Tab.~\ref{tab:extra_md}.

\begin{table}[h]
	\centering
	\vskip -0.05in
	\caption{Performance of misclassification detection for more datasets and networks. AURC and E-AURC values are multiplied
		by $10^3$. Other values are percentages. The best results are boldfaced for highlighting.}
	\label{tab:extra_md}
	\vskip -0.05in
	\renewcommand{\arraystretch}{1.15}
	\renewcommand\tabcolsep{5.0pt}
	\scalebox{0.70}{
		\begin{tabular}{cccccccccccc}
			\toprule[1.4pt]
			\multicolumn{1}{c}{\multirow{2}{*}{Model}} & \multirow{2}{*}{Method}   & \multicolumn{5}{c}{CIFAR-10}                                                                                                & \multicolumn{5}{c}{CIFAR-100}                                                                                                               \\  \cmidrule(r){3-7}  \cmidrule(r){8-12}
			\multicolumn{1}{c}{}      &   & \multicolumn{1}{c}{AUROC $\uparrow$} & \multicolumn{1}{c}{AURC $\downarrow$} & \multicolumn{1}{c}{E-AURC $\downarrow$} & \multicolumn{1}{c}{AUPR-Err $\uparrow$} & FPR95 $\downarrow$ & \multicolumn{1}{c}{AUROC $\uparrow$} & \multicolumn{1}{c}{AURC $\downarrow$} & \multicolumn{1}{c}{E-AURC $\downarrow$} & \multicolumn{1}{c}{AUPR-Err $\uparrow$} & FPR95 $\downarrow$  \\ \cmidrule(r){1-2}  \cmidrule(r){3-7}  \cmidrule(r){8-12}
			\multirow{2}{*}{Vgg-16}    & MSP  &  91.04 & 10.36 & 7.97 & 44.63  &  45.60   &    85.91 & 91.12 & 47.21 & 67.70  &  66.06
			\\
			& \cellcolor{gray!30} {$\quad$ + AoP} & \cellcolor{gray!30} \cellcolor{gray!30}\textbf{91.47} & \cellcolor{gray!30}\textbf{9.45} & \cellcolor{gray!30}\textbf{7.31} & \cellcolor{gray!30}\textbf{45.25} & \cellcolor{gray!30}\textbf{41.40}      &  \cellcolor{gray!30}\textbf{86.51} & \cellcolor{gray!30}\textbf{86.23} & \cellcolor{gray!30}\textbf{42.74} & \cellcolor{gray!30}\textbf{67.99} & \cellcolor{gray!30}\textbf{64.93}                     
			\\ \cmidrule(r){1-2}  \cmidrule(r){3-7}  \cmidrule(r){8-12}
			\multirow{2}{*}{ResNet-18}    & MSP   & 93.41   & 6.06  & 4.32 & \textbf{45.88} & 39.02    &   86.03   & 84.75  & 43.31 & 66.29 & 66.53  \\ 
			&  \cellcolor{gray!30}  {$\quad$ + AoP} &    \cellcolor{gray!30}\textbf{93.77}   & \cellcolor{gray!30}\textbf{5.52}  & \cellcolor{gray!30}\textbf{3.90} & \cellcolor{gray!30}{45.59}   & \cellcolor{gray!30}\textbf{37.17}   &  \cellcolor{gray!30}\textbf{86.25}   & \cellcolor{gray!30}\textbf{81.76}  & \cellcolor{gray!30}\textbf{42.06} & \cellcolor{gray!30}\textbf{66.42} & \cellcolor{gray!30}\textbf{65.98}     
			\\ \toprule[1.4pt]
	\end{tabular}}
\end{table}

\section{Discussions}
\label{sec:future_work}

In this paper, we show that the performance of OOD detection suffers from instability and overfitting in the training stage. \textit{Our findings reveal that the training dynamics provide an interesting research perspective ignored by previous studies in the field of OOD detection.} The method is this paper is very simple and effective.

We use model averaging to stabilize the performance of OOD detection in training. The realization is convenient and has almost no computational overhead. We adopt LTH~\cite{frankle_2018_LTH,frankle_2020_wrewind} as the pruning method, which is a very well-known method for post-training pruning. LTH has detailed implementation codes and is easy to reproduce the results. Besides, its realization is consistent with our theoretical motivation from LASSO, which could certify our theoretical results. We have also verified that various during-training
pruning methods like GMP~\cite{zhu2017GMP}, RigL~\cite{evci2020rigl}, and GraNet~\cite{liu2021GraNet} could improve the performance of OOD detection, which are often more efficient. Various pruning techniques could achieve improvement in OOD detection rather than just LTH. Comprehensive experiments in the paper verify the effectiveness of pruning in OOD detection.

\end{document}